\PassOptionsToPackage{table}{xcolor}
\documentclass{article}
\usepackage{iclr2027_conference,times}
\usepackage{hyperref}
\usepackage{url}

\usepackage{booktabs}
\usepackage{amsmath,amssymb}
\usepackage{subcaption}
\usepackage{xcolor}
\usepackage{multirow}
\usepackage{makecell}
\usepackage{algorithm}
\usepackage{soul}
\usepackage{enumitem}   % itemize spacing options used by the contribution list
\usepackage{algpseudocode}
\usepackage{microtype}
\usepackage{graphicx}
\usepackage{pifont}
\setstcolor{red}

\title{SHield: Watermarking 4D Gaussian Splatting via Spherical Harmonic Coefficient Perturbation}

\author{%
\makebox[\textwidth][c]{\textbf{Kunal Bhosikar}$^{1, 3}$\hspace{1.2em}\textbf{Prajas Wadekar}$^{1}$\hspace{1.2em}\textbf{Venkata Sai Pranav Bachina}$^{1}$}\\[0.25em]
\makebox[\textwidth][c]{\textbf{Ankit Gangwal}$^{1}$\hspace{1.2em}\textbf{Charu Sharma}$^{1}$\hspace{1.2em}\textbf{Nicole Meng}$^{2}$\hspace{1.2em}\textbf{Yingjie Lao}$^{2}$}\\[0.45em]
\makebox[\textwidth][c]{\normalsize $^{1}$IIIT Hyderabad\hspace{1.6em}$^{2}$Tufts University\hspace{1.6em}$^{3}$Miko}\\[0.2em]
\makebox[\textwidth][c]{\footnotesize\texttt{\{kunal.bhosikar, prajas.wadekar\}@research.iiit.ac.in,} \texttt{bachina.pranav@gmail.com}}\\[0.1em]
\makebox[\textwidth][c]{\footnotesize\texttt{\{gangwal, charu.sharma\}@iiit.ac.in, \{ziyi.meng, yingjie.lao\}@tufts.edu}}
}

\iclrfinalcopy % Uncomment for camera-ready version, but NOT for submission.
\begin{document}

\maketitle
\lhead{Preprint. Under review.}
% ─────────────────────────────────────────────────────────────────────────────
% \noindent
% \begin{minipage}{\textwidth}
%     \centering
%     \includegraphics[width=\textwidth]{Figures/SHield_final_teaser_figure.drawio.png}
%   % \captionof{figure}{%
%   %   \textbf{SHield verifies a 4DGS model file, not rendered video.} The owner writes an RGB payload
%   %   into the canonical SH coefficients of key-selected Gaussians and registers a private key.
%   %   To verify a suspect, possibly post-processed copy of the file, the owner re-associates the key with
%   %   the file's Gaussians, recovers the payload from the SH parameters in closed form and accepts if the
%   %   correlation with the registered payload reaches $\tau$. Example: \textsc{vrig-chicken}, wm1, $39.9$~dB WM-PSNR from the distributed file and
%   %   $34.7$~dB after whole-file INT8.}
%     \captionof{figure}{\textbf{SHield verifies a model file, not rendered video.} The owner writes a zero-mean payload
%     into the canonical SH coefficients of key-selected Gaussians; an adversary post-processes the redistributed
%     \texttt{.ply}; verification re-associates the registered key by Gaussian centres and recovers the payload in
%     closed form, with no rendering and no learned decoder. \hl{move to the 2nd page? Somehow the style and fonts look different from Figure 2}}
%     \label{fig:teaser}
% \end{minipage}
% \vspace{0.4em}

% ─────────────────────────────────────────────────────────────────────────────
\begin{abstract}
4D Gaussian Splatting (4DGS) models are computationally expensive to train but distributed as easily copied parameter files. 
Existing watermarks protect these assets by fine-tuning the scene against a neural decoder to extract bit strings from rendered video, a heavy process vulnerable to the temporal dynamics of dynamic scenes. 
However, if the threat model targets the redistributed file itself, verification can be vastly simplified. 
Because the canonical Spherical-Harmonic (SH) coefficients of 4DGS are time- and view-invariant, parameter-space perturbations can be recovered via deterministic arithmetic. 
We propose \textbf{SHield}, the first non-blind, decoder-free watermarking approach that natively embeds full-resolution RGB images into 4DGS files. 
SHield tiles a zero-mean payload across the SH coefficients of a key-selected Gaussian subset, applying a brief anchored optimization to preserve visual fidelity. 
To verify ownership, it geometrically re-associates the key using canonical Gaussian centres to survive file tampering and recovers the payload in closed form. 
Across 30 scenes (D-NeRF, HyperNeRF, DyNeRF), SHield guarantees near-perfect payload recovery (Pearson $r \geq 0.991$) on all 150 tested models with a mean rendering penalty of $\leq 0.13$~dB. 
Held-out evaluation yields perfect detection on 400 true trials and zero false positives across $1{,}600$ wrong-subset null trials. 
We also strictly bound the method's robustness: SHield survives INT8 quantization and $30\%$ opacity pruning for the image payload, and on aggressive SH-degree-1 truncation bit payloads are still decodable. 
% Furthermore, we demonstrate that an adaptive adversary can photometrically re-optimize the model to strip the watermark at a rendering cost of under $1$~dB.
\end{abstract}

\begin{figure}[h]
    \centering
    \includegraphics[width=\linewidth]{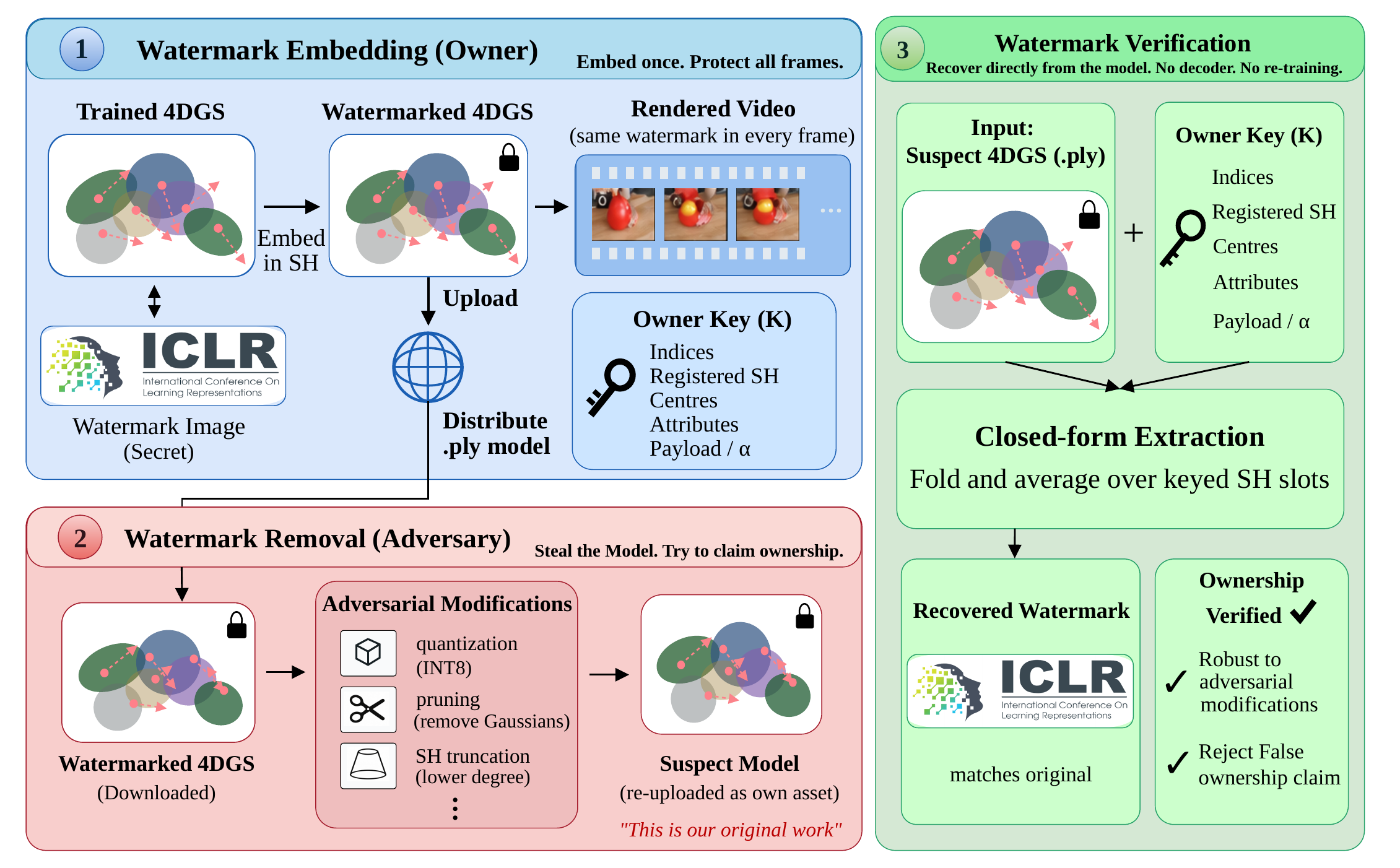}
    \caption{\textbf{File-centric ownership verification with SHield.} 
    The owner embeds a payload into the SH coefficients of key-selected Gaussians.
    Under adversarial file degradation, verification geometrically re-associates the key via canonical centres, enabling closed-form extraction directly from the parameters without learned decoders or rendering loops.}
    \label{fig:teaser}
\end{figure}

% ─────────────────────────────────────────────────────────────────────────────
\section{Introduction}
\label{sec:intro}

4D Gaussian Splatting (4DGS)~\citep{wu20244d,yang2024real} achieves real-time dynamic scene rendering by displacing explicit Gaussians via a learned deformation field~\citep{kerbl20233d}. 
Because training these models requires immense data and compute, the resulting parameter files are valuable intellectual property. 
However, they are also trivially easy to copy and redistribute, creating an urgent need for robust ownership verification.

Most Gaussian splatting watermarks operate in the render domain: they fine-tune the scene against a learned decoder until a short bit string survives the rendering process~\citep{chen2025guardsplat,jang20253dgsw,huang2024gaussianmarker}. 
In 4DGS, this introduces a severe temporal challenge. 
A perturbation that is invisible at one timestamp may become glaringly obvious at the next timestamp as the Gaussians move, this phenomenon is known as ``FVD collapse''~\citep{zhou20264dgsw}.
Existing methods counter this with heavy, motion-specific machinery. 
For example, Mark4D~\citep{lee2026mark4d} uses trajectory-aligned offsets and a video-text decoder, while 4D-GSW relies on curvature-based gating. 
This forces thousands of fine-tuning iterations (e.g., $4{,}000$ steps for Mark4D) just to embed a few dozen bits. 
While this complexity is necessary if the watermark must be extracted from a rendered video, it is vastly over-engineered if the adversary is simply redistributing the underlying model file.

Our approach leverages the fundamental architecture of 4DGS appearance storage. 
Each Gaussian maintains a single, canonical set of spherical-harmonic (SH) coefficients that the deformation field queries across all timestamps and viewpoints. 
By writing a perturbation directly into these canonical coefficients, we create a time-invariant watermark. 
Shifting the verification target from the rendered video back to the redistributed file itself eliminates the need for learned decoders and rendering loops, allowing extraction via deterministic arithmetic while completely sidestepping temporal fragility.

We introduce \textbf{SHield} (Figure~\ref{fig:teaser}), the first decoder-free watermarking framework that natively embeds full-resolution RGB images into 4DGS parameter files and recovers them in closed form. 
SHield operates as a non-blind watermark, meaning verification requires the owner's private key rather than the full original \texttt{.ply} file. 
The embedding process tiles a zero-mean payload across the SH coefficients of a key-selected subset of Gaussians. 
Because naive tiling ignores whether a Gaussian is actually visible, SHield employs a brief \(T\)-step anchored optimization. 
This step adapts the perturbation to minimize visual rendering error while tightly anchoring each coefficient to its target value. 
During verification, the algorithm geometrically re-associates the owner's key with the suspect file using canonical centres (resolving ties via unchanged attributes), subtracts the stored reference coefficients, and averages the redundant copies. 
Crucially, SHield modifies only the SH coefficients; the scene geometry and deformation network remain strictly frozen.
Our main contributions can be summarized as follows:

\begin{itemize}[leftmargin=1em,itemsep=1pt,topsep=0pt,parsep=0pt]
    \item \textbf{First full-resolution image watermark for 4DGS.} 
    To our knowledge, SHield is the first method to embed and recover a full-resolution RGB image directly from a 4DGS model file. 
    Prior dynamic scene watermarks rely on learned decoders without reporting image resolution~\citep{liu2025hideinmotion}, and full-resolution steganography has previously been restricted to static 3DGS~\citep{meng2026whispersplat}.
    \item \textbf{Decoder-free, non-blind verification.} 
    Extraction is reduced to closed-form arithmetic on canonical SH coefficients. 
    By introducing centre-based re-association and a coprime stride, the watermark survives adversarial row reordering, opacity pruning, and SH-degree truncation.
    \item \textbf{Anchored optimization.} 
    We propose a brief optimization strategy that adapts the tiled perturbation to the renderer's sensitivity. 
    This reduces the visual footprint of plain tiling by \(1.6\times\), \(5.6\times\), and \(3.0\times\) across the three benchmark datasets, trading a modest drop in WM-PSNR for significantly higher rendering fidelity.
    \item \textbf{Rigorous robustness evaluation.} 
    We evaluate SHield across $150$ models from $30$ scenes~\citep{pumarola2021d,park2021hypernerf,li2022neural}. 
    We thoroughly map both its capabilities and operational limits, testing against non-adaptive post-processing (e.g., pruning, quantization) as well as adaptive attacks including targeted fine-tuning, duplication, and rigid transformations.
\end{itemize}
% ─────────────────────────────────────────────────────────────────────────────
\section{Related Work}
\label{sec:related}

\paragraph{Image and video watermarking.}
Classical spread-spectrum watermarking redundantly embeds messages across host coefficients, verifying authenticity by correlating the recovered signal against a registered reference. 
While blind detectors operate solely on suspect content, informed (non-blind) detectors leverage auxiliary owner-side data~\citep{cox2007digital}. 
Modern deep watermarking subsequently replaced hand-crafted embeddings with neural encoder--decoder pairs trained against synthetic distortions for images~\citep{zhu2018hidden,tancik2020stegastamp,ma2022cin,fernandez2022ssl} and video~\citep{zhang2019rivagan,fernandez2024videoseal}. 
SHield adapts this classical informed-correlation approach, shifting the host domain from pixel space to the internal parameter file of a 4DGS asset.

\paragraph{Radiance fields and static 3DGS.}
Neural asset protection initially targeted continuous coordinate fields, where CopyRNeRF~\citep{luo2023copyrnerf} and StegaNeRF~\citep{li2023steganerf} fine-tune Neural Radiance Fields to allow message recovery from rendered novel views. 
A similar render-centric paradigm dominates explicit 3D Gaussian Splatting: GuardSplat~\citep{chen2025guardsplat} optimizes spherical-harmonic (SH) offsets against a CLIP-guided decoder, 3D-GSW~\citep{jang20253dgsw} incorporates frequency-guided densification, and GaussianMarker~\citep{huang2024gaussianmarker} anchors watermarks using per-Gaussian uncertainty. 
To withstand compression, CompMarkGS~\citep{in2025compmarkgs} and Robust3DGSW~\citep{wang2026robust3dgsw} integrate quantization-aware training into the pipeline. 
Broader steganographic frameworks extend these ideas to hide complex content, ranging from GS-Hider's~\citep{zhang2024gshider} coupled feature substitution and KeySS's~\citep{ren2026keyss} optimized key-secured decoders to Splats in Splats~\citep{guo2026splats,guo2026splatspp} encrypting secondary scenes into SH attributes. 
Notably, WhisperSplat~\citep{meng2026whispersplat} avoids model modification entirely, learning an SH noise key that renders a secret high-resolution image from a single viewpoint without an explicit decoder.

\paragraph{Dynamic Gaussians.}
Adapting watermarks to 4DGS requires marks to survive continuous motion. 
Mark4D~\citep{lee2026mark4d} learns time-dependent offsets decoded via video encoders, while 4D-GSW~\citep{zhou20264dgsw} gates optimization gradients by trajectory curvature. 
Guard4D~\citep{shi2026guard4d} also utilizes SH coefficients with frozen geometry, yet still relies on text-supervised decoders operating on rendered clips. 
Hide-in-Motion~\citep{liu2025hideinmotion} embeds multimodal payloads into composite attributes but requires a private U-Net decoder and omits post-processing evaluation. 
Splats in Splats++~\cite{guo2026splatspp} similarly depends on a render-and-recover framework.

\paragraph{Where SHield differs.}
While prior approaches bind verification to rendering loops or neural decoders, SHield treats the redistributed model file itself as the verification object. 
Eliminating this overhead enables full-resolution image recovery via direct, closed-form arithmetic. 
This file-centric reliability requires a non-blind design where the owner matches a registered private key.

% ─────────────────────────────────────────────────────────────────────────────
\section{Method}
\label{sec:method}

\subsection{Background and Threat Model}
\label{sec:background}

3DGS~\citep{kerbl20233d} parameterizes a scene using \(N\) Gaussians \(\mathcal{G}=\{(\boldsymbol{\mu}_i,\mathbf{q}_i,\mathbf{s}_i,o_i,\mathbf{h}_i)\}_{i=1}^{N}\), defined by their centre, rotation, scale, opacity, and canonical spherical harmonic (SH) coefficients \(\mathbf{h}_i\in\mathbb{R}^{C_\text{tot}\times 3}\) (\(C_\text{tot}{=}16\) for degree $3$). 
SHield exclusively modifies the SH coefficients, explicitly preserving the geometric and opacity attributes \(\mathbf{a}_i=(o_i,\mathbf{q}_i,\mathbf{s}_i)\) to serve as invariant geometric anchors for key re-association. 
In 4DGS~\citep{wu20244d}, a learned deformation network predicts time-dependent geometric offsets. 
Crucially, the stored SH coefficients \(\mathbf{h}_i\) remain canonical and time-invariant; even in architectures like DyNeRF that inject time-dependent SH offsets during rendering, the underlying file parameters remain completely unmodified.

The owner embeds an RGB image \(\mathbf{W}\in[0,1]^{H_0\times W_0\times 3}\) or an \(L\)-bit message \(\mathbf{m}\in\{0,1\}^{L}\) into a trained 4DGS model, distributes the watermarked file, and retains a private key \(\mathcal{K}\). 
For non-blind extraction, \(\mathcal{K}\) stores the selected Gaussian indices, original SH coefficients, canonical centres, static tie-break attributes, and payload metadata. 
Ownership is verified via Pearson correlation \(r(\hat{\mathbf{W}},\mathbf{W}) \geq \tau\) for images, or direct decoding accuracy for bit payloads.
We assume the owner registers \(\mathcal{K}\) and the payload prior to redistribution to establish provenance.

We assume a white-box adversary holding the \texttt{.ply} file who can modify any parameter and knows the algorithm, but lacks the private key \(\mathcal{K}\) (Kerckhoffs' principle~\citep{Kerckhoffs1883}). 
We evaluate non-adaptive post-processing (e.g., SH quantization, reordering, pruning, truncation, and noise) alongside adaptive attacks targeting model re-optimization or the key matcher (\S~\ref{sec:attack}).
We assume the redistributed file retains its float32 canonical centres in the original coordinate frame.

\begin{figure}
    \centering
    \includegraphics[width=\linewidth]{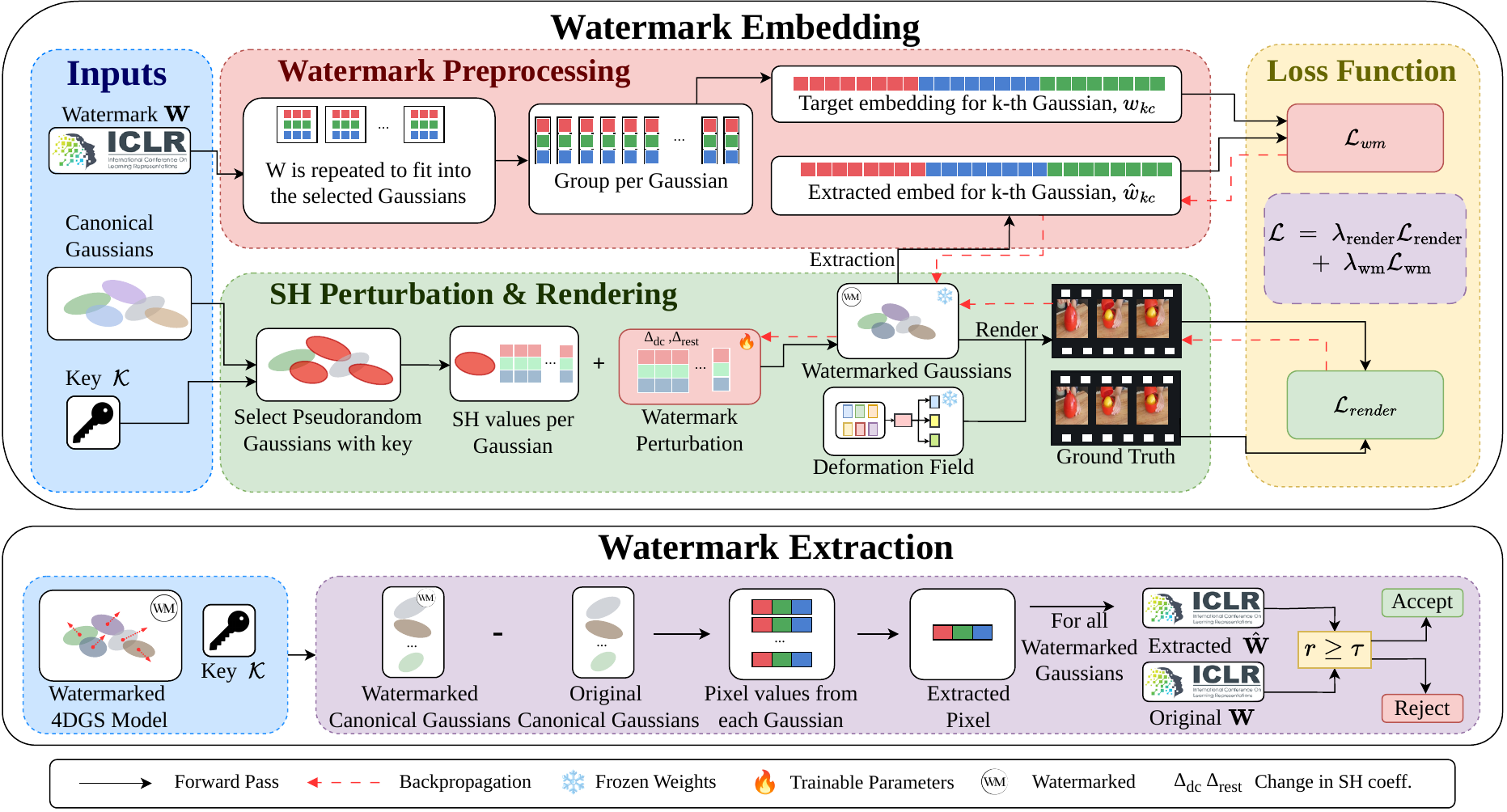}
    \caption{\textbf{Overview of SHield.} 
    \emph{(a) Embedding:} A zero-mean payload is tiled across the SH slots of key-selected Gaussians (red). 
    Learnable offsets are applied to these SH coefficients while the scene geometry, unselected Gaussians, and deformation network remain strictly frozen (green). 
    A joint loss anchors the offsets to the tiled target while preserving rendering fidelity (yellow). \emph{(b) Verification:} Centre-based re-association locates the target Gaussians even if the file is reordered or pruned. 
    Subtracting the stored reference SH cleanly isolates the perturbation, and the redundant copies are averaged to reconstruct the payload \(\hat{\mathbf{W}}\).
    Ownership is verified if the correlation \(r \geq \tau\).}
\label{fig:overview}
\end{figure}

\subsection{Watermark Embedding}
\label{sec:embedding}
SHield embeds the watermark into the canonical SH coefficients (Figure~\ref{fig:overview}). 
This provides two advantages: a single time-invariant perturbation propagates to all timestamps, and SH coefficients offer a robust carrier with minimal visual sensitivity. 
At matched relative strength, modifying spatial positions degrades rendering PSNR by \(\sim\)$50$~dB more than modifying higher-order SH bands (Appendix~\ref{app:carrier}). 
The mathematical protocol is detailed below (full pseudocode in Algorithm~\ref{alg:embed}).

% \paragraph{Tiled SH embedding.}
% The private key \(\mathcal{K}\) seeds a pseudorandom generator to select a subset of \(K\) Gaussians without replacement, forming the ordered index set \(\mathcal{S} = (k_1 < \cdots < k_K) \subset [N]\). 
% The subset size \(K\) dynamically scales to ensure at least \(n_t^{\min}\) complete payload copies (tiles), while strictly bounding the modified population between a minimum (\(\rho_\text{min}\)) and maximum (\(\rho_\text{max}\)) fraction of the total Gaussians:
% \begin{equation}
%     K = \min\!\Bigl(\max\!\bigl(\lfloor\rho_\text{min} N\rfloor,\;\lceil n_t^{\min} n_\text{wm} / C_\text{tot}\rceil\bigr),\;\lfloor\rho_\text{max} N\rfloor\Bigr),
%     \label{eq:kselect}
% \end{equation}
% where \(n_\text{wm}=H_0W_0\) and \(n_t = \lfloor K C_\text{tot} / n_\text{wm} \rfloor\) represents the number of complete tiles.
% For each colour channel \(c\), the payload is zero-meaned to \(\mathbf{z}_c\) (storing the mean \(\bar{w}_c\) in \(\mathcal{K}\)) and tiled sequentially across the \(K C_\text{tot}\) available slots to construct the target perturbation matrix \(\mathbf{E}_c\). 
% The watermarked coefficients are initialized as \(\mathbf{h}_{k_j,c}^\text{wm} = \mathbf{h}_{k_j,c} + \alpha_0\,\mathbf{E}_c[j,:]\).

\paragraph{Tiled SH embedding.}
The key \(\mathcal{K}\) pseudorandomly selects an ordered subset of \(K\) Gaussians \(\mathcal{S} = (k_1 < \cdots < k_K) \subset [N]\). 
\(K\) scales to ensure at least \(n_t^{\min}\) complete payload copies (tiles), strictly bounded between population fractions \(\rho_\text{min}\) and \(\rho_\text{max}\):
\begin{equation}
    K = \min\!\Bigl(\max\!\bigl(\lfloor\rho_\text{min} N\rfloor,\;\lceil n_t^{\min} n_\text{wm} / C_\text{tot}\rceil\bigr),\;\lfloor\rho_\text{max} N\rfloor\Bigr),
    \label{eq:kselect}
\end{equation}
where \(n_\text{wm}=H_0W_0\) and \(n_t = \lfloor K C_\text{tot} / n_\text{wm} \rfloor\) is the number of complete tiles. 
For each channel \(c\), the zero-meaned payload \(\mathbf{z}_c\) (storing mean \(\bar{w}_c\) in \(\mathcal{K}\)) is tiled across the \(K C_\text{tot}\) slots to form the target perturbation matrix \(\mathbf{E}_c\). 
Coefficients are initialized as \(\mathbf{h}_{k_j,c}^\text{wm} = \mathbf{h}_{k_j,c} + \alpha_0\,\mathbf{E}_c[j,:]\).

For \(L\)-bit messages, the payload is whitened via a key-derived sequence and mapped to \(\pm1\). 
Because \(C_\text{tot}{=}16\) cleanly divides standard message lengths (\(L \in \{32, 48, 64\}\)), plain tiling would destructively concentrate all copies of a specific bit onto a single SH coefficient index. 
To prevent complete message erasure during SH-degree truncation, SHield shifts each tile \(t\) by \(t\sigma\), where the stride \(\sigma\) is strictly coprime to both \(L\) and \(C_\text{tot}\) (Appendix~\ref{app:stride}).

\paragraph{Reducing the rendering cost.}
\label{sec:shield}
Plain tiling applies a uniform amplitude regardless of a Gaussian's visibility; thus, a large \(\alpha_0\) degrades visual quality while a small \(\alpha_0\) risks quantization vulnerability. 
SHield resolves this by fixing \(\alpha_0\) at a robust level and optimizing learnable SH offsets (initialized at \(\alpha_0\mathbf{E}\)) for \(T\) steps across \(n_v\) training views via the frozen deformation network, minimizing:
\begin{align}
    \mathcal{L} &= \lambda_\text{render}\,\mathcal{L}_\text{render} + \lambda_\text{wm}\,\mathcal{L}_\text{wm}, \label{eq:loss} \\
    \mathcal{L}_\text{render} &= \ell_1(\hat{I}, I) + \lambda_\text{ssim}(1 - \text{SSIM}(\hat{I}, I)), \label{eq:lrender} \\
    \mathcal{L}_\text{wm} &= \tfrac{1}{3} \textstyle\sum_{c}\text{MSE}\bigl(\mathbf{e}_c[1{:}n_t n_\text{wm}],\; \alpha_0\cdot\text{tile}(\mathbf{z}_c,\,n_t)\bigr), \label{eq:lwm}
\end{align}
where \(\hat{I}\) and \(I\) are the watermarked and ground-truth renders. 
\(\mathcal{L}_\text{render}\) adapts offsets to the scene's topology, while \(\mathcal{L}_\text{wm}\) anchors the \(n_t n_\text{wm}\) active slots to their tiled targets. 
This minor departure from plain tiling (averaging just \(1.8\%\) on D-NeRF) drastically improves visual fidelity.

\begin{algorithm}
    \footnotesize
    \caption{Verification (images)}
    \label{alg:verify}
    \begin{algorithmic}[1]
        \Require suspect file (centres $\tilde{\boldsymbol{\mu}}$, attributes $\tilde{\mathbf{a}}$, SH $\tilde{\mathbf{h}}$); key $\mathcal{K}$; $\tau$
        \Ensure decision, confidence $r$
        \Statex \Comment{\textit{Phase 1: Geometric Re-association}}
        \State $\pi \gets \textsc{ReAssociate}(\boldsymbol{\mu}[\mathcal{S}], \mathbf{a}[\mathcal{S}], \tilde{\boldsymbol{\mu}}, \tilde{\mathbf{a}})$
        \State $\mathbf{E}_k \gets (\tilde{\mathbf{h}}_{\pi_k} - \mathbf{h}_\text{orig}[\mathcal{S}]_k)/\alpha_0$ \Comment{0 if $\pi_k{=}\varnothing$}
        \Statex \Comment{\textit{Phase 2: Fold-and-Average}}
        \For{$c \in \{R,G,B\}$}
          \State $\mathbf{e}_c^{(t)} \gets$ tile $t$ of $\textsc{Flatten}(\mathbf{E}[:,:,c])$
          \State $\hat{\mathbf{z}}_c \gets \frac{1}{n_t}\sum_{t}\textsc{Rot}\bigl(\mathbf{e}_c^{(t)}, {+}t\sigma\bigr)$ \Comment{Eq.~\ref{eq:foldavg}; undoes the ${-}t\sigma$ embedding shift, $\sigma{=}0$ for images}
          \State $\hat{\mathbf{W}}_c \gets \text{clip}(\hat{\mathbf{z}}_c + \overline{\mathbf{W}_c},\,0,\,1)$
        \EndFor
        \Statex \Comment{\textit{Phase 3: Statistical Verification}}
        \State $r \gets \text{Pearson}(\hat{\mathbf{W}}, \mathbf{W})$
        \State \Return $r \geq \tau$ \Comment{$\tau{=}0.6$}
    \end{algorithmic}
\end{algorithm}

\subsection{Verification Interface}
Verification proceeds completely in the parameter space without requiring any rendering loops or learned decoders (Algorithm~\ref{alg:verify}).

\paragraph{Key re-association.}
\label{sec:reassoc}
Because adversarial redistribution often involves Gaussian pruning or row permutation (Figure~\ref{fig:overview}b), the verification key cannot safely rely on static matrix indices. 
Instead, SHield geometrically re-associates each key entry by matching it one-to-one with a suspect Gaussian whose canonical centre falls within a tight tolerance (\(\epsilon = 0.01\,s_\text{nn}\)), where \(s_\text{nn}\) is the original model's median nearest-neighbour spacing. 
If multiple candidates fall within \(\epsilon\), the static attributes (opacity, scale, and rotation) act as strict tie-breakers.
Any entry lacking a unique, valid candidate is securely treated as missing rather than guessed, preventing geometric noise from corrupting the extraction array.

\paragraph{Fold-and-average extraction.}
\label{sec:extract}
Following re-association, the perturbation for key entry \(k\) is isolated via \(\mathbf{E}_c^\text{rec}[k] = (\tilde{\mathbf{h}}_{\pi_k,c} - \mathbf{h}_{\mathcal{S}_k,c})/\alpha_0\) (where \(\pi_k\) is the matched Gaussian), then flattened and averaged across all \(n_t\) tiles:
\begin{equation}
  \hat{\mathbf{z}}_c = \frac{1}{n_t} \sum_{j=0}^{n_t-1} \mathbf{e}_c[j\,n_\text{wm} : (j{+}1)\,n_\text{wm}].
  \label{eq:foldavg}
\end{equation}
The payload is reconstructed as \(\hat{\mathbf{w}}_c = \text{clip}(\hat{\mathbf{z}}_c + \bar{w}_c, 0, 1)\) (bit messages are first inverse-shifted by \(t\sigma\) and un-whitened). 
Because subtraction exactly cancels the scene coefficients, errors on clean files stem only from optimizer departures, clipping, and 8-bit rounding.
Unmatched or pruned Gaussians contribute zero, gracefully attenuating the payload rather than corrupting the data stream.

% ─────────────────────────────────────────────────────────────────────────────
\section{Experiments}
\label{sec:experiments}

\subsection{Experimental Setup}
We evaluate standard 4DGS~\citep{wu20244d} across $30$ scenes ($7$ D-NeRF, $17$ HyperNeRF, $6$ DyNeRF).
Embedding five RGB payloads ($50$k–$217$k pixels) yields $150$ watermarked models.
Rendering quality (PSNR, SSIM~\citep{wang2004image}, LPIPS~\citep{zhang2018unreasonable}) is measured against ground truth on $20$ test views, and extraction via WM-PSNR and Pearson \(r\). 
Because parameter-space extraction shares no protocol with render-domain methods, our primary baseline is SHield's plain-tiling initialization (at matched amplitude and selection) to strictly isolate the impact of anchored optimization.
Implementation details are in Appendix~\ref{app:impl}.

\begin{table}
\centering
\caption{\textbf{Rendering cost and extraction} ($\alpha_0{=}0.1$, five payloads per scene, $20$ test views).
$\Delta$ values are against the unwatermarked model, with the worst scene bracketed; min~$r$ is the worst payload
recovery over models. The comparison against plain tiling at the same $\alpha_0$ and selection is in
Appendix~\ref{app:ablation}.}
\label{tab:main_results}
\setlength{\tabcolsep}{3pt}
\footnotesize
{\begin{tabular}{l r c c c r r}
\toprule
Dataset (models) & PSNR$_\text{clean}$ & $\Delta$PSNR [worst] & \makecell{SSIM\\($\Delta$)} & \makecell{LPIPS\\($\Delta$)} & WM-PSNR & min $r$ \\
\midrule
D-NeRF (35)~\citep{pumarola2021d}    & 35.31 & $-$0.048 [$-$0.113] & \makecell{0.9842\\($-$0.0004)} & \makecell{0.0164\\($+$0.0002)} & 44.77 & 0.9990 \\
HyperNeRF (85)~\citep{park2021hypernerf} & 26.92 & $-$0.047 [$-$0.150] & \makecell{0.7761\\($-$0.0012)} & \makecell{0.2949\\($+$0.0009)} & 36.27 & 0.9910 \\
DyNeRF (30)~\citep{li2022neural}    & 31.19 & $-$0.127 [$-$0.314] & \makecell{0.9353\\($-$0.0011)} & \makecell{0.1099\\($+$0.0012)} & 36.59 & 0.9952 \\
\bottomrule
\end{tabular}}
\end{table}

\begin{figure}[h]
    \centering
    \includegraphics[width=\textwidth]{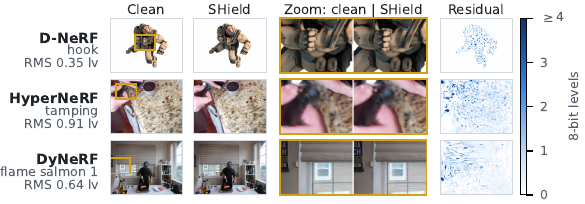} % printed type ~8.1 pt, same as Figure 4
    \caption{\textbf{Rendering residual} ($\alpha_0{=}0.1$, wm1): each dataset's median-residual scene at its median
    test view, to avoid cherry-picking; per-pixel RMS difference to the clean render in 8-bit levels (lv), one shared
    scale, zoom box at the largest local residual.}
\label{fig:diffmaps}
\end{figure}

\subsection{Rendering Quality and Extraction}
\label{sec:extraction}
Mean rendering penalties are just $0.048$, $0.047$, and $0.127$ dB on D-NeRF, HyperNeRF, and DyNeRF, with no scene degrading $>0.314$ dB and extraction remaining near-perfect (\(r \geq 0.991\)) across all $150$ models (Table~\ref{tab:main_results}).
Compared to plain tiling, anchored optimization reduces rendering costs by \(1.6\times\), \(5.6\times\), and \(3.0\times\) (improving 149/150 pairs).
This trades raw extraction fidelity for visual quality (e.g., DyNeRF WM-PSNR drops from $56.2$ to $36.6$ dB), yet correlation remains exceptionally robust (\(r \geq 0.995\)).

For wm1, the per-view RMS residual across $830$ test views is \(\leq 1.85\) 8-bit levels and spatially diffuse (Figure~\ref{fig:diffmaps}). 
The SH distribution shifts by a 3D-Sinkhorn distance~\citep{ren2026keyss} of \(<0.02\), near the $0.008$ quantization floor.
Parameter-space operation guarantees view-independent extraction: while a 2D baseline applied to training frames fails on novel views (\(r=0.004\)), SHield reliably reads \(r=0.998\) directly from the file (Appendix~\ref{app:2d_baseline}).

\subsection{Verification and Key Re-association}
\label{sec:verification}

\begin{figure}[h]
  \centering
  \includegraphics[width=\textwidth]{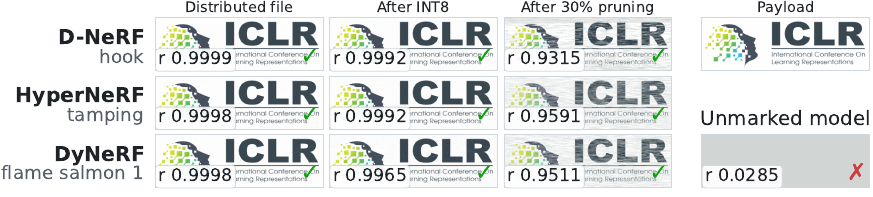}
  \caption{\textbf{Payloads recovered from the model file}, for the scenes of Figure~\ref{fig:diffmaps}: as
  distributed, after whole-file SH INT8, and after $30\%$ opacity pruning, with Pearson $r$ against $\tau{=}0.6$.
  On an unmarked model the recovery collapses to a flat block of the payload's per-channel mean colour, whose
  residual correlation is payload-dependent.}
  \label{fig:qualitative}
\end{figure}

\paragraph{Held-out verification.}
$\tau=0.6$ was chosen by an earlier in-sample analysis of all $150$ models and then evaluated unchanged on a
scene-disjoint split of $14$ calibration and $16$ held-out scenes. On the held-out scenes ($80$ models) SHield accepted all $640$ true trials -- the distributed file, whole-file INT8,
$30\%$ opacity pruning, and $400$ fresh trials under $30\%$ random deletion, row permutation and key re-association
(minimum $r=0.815$) -- with no false positive in $2{,}320$ null trials of three kinds: another owner's payload on the
same scene ($320$, maximum $r=0.317$), a wrong subset of Gaussians ($1{,}920$, maximum $0.498$) and an unmarked model
($80$, maximum $0.368$; undefined $r$ counts as a non-detection). Nominal one-sided $95\%$ Clopper--Pearson bounds, treating
models or scenes as independent, are $3.68\%$ and $17.1\%$; the independence is imperfect, so these are not
deployment-level rates (Appendix~\ref{app:verification}). The in-sample choice of $\tau$ does not carry the result: every decision is
identical for any $\tau \in (0.498, 0.815]$, and a threshold from the calibration scenes alone ($0.657$) reproduces
them exactly. A centred statistic, which removes the payload-dependent residual of
Figure~\ref{fig:qualitative}, makes the same held-out decisions at a calibration-only threshold ($\tau_c=0.5834$;
$240/240$ accepted, no false positive in $720$) but does not repair the SH-truncation behaviour below, so we keep the
uncentred one.

\paragraph{Re-association.}
Across all $192$ models, re-association reproduces extraction exactly on unmodified and permuted files, and after
$30\%$ lowest-opacity or $20\%$ random deletion every surviving marked Gaussian finds its correct row ($0.008\%$ and
$0.116\%$ of removed entries falsely matched). Geometric noise is the weak point: all marked Gaussians match under
isotropic centre noise up to $0.001\,s_\text{nn}$ but only $19.9\%$ at $0.01\,s_\text{nn}$, where image verification
fails on $103$ of $150$ models, and a float16 cast leaves $3.75\%$ matched and fails $135$ of $150$, though bit
payloads still decode (Appendix~\ref{app:reassoc}).

\begin{table}
    \centering
    \small
    \caption{\textbf{Post-processing of the model file} ($\alpha_0{=}0.1$, all $150$ image-payload models): mean
    WM-PSNR, mean $r$ and passes at $\tau{=}0.6$.}
\label{tab:robustness}
    \setlength{\tabcolsep}{4.5pt}
    \begin{tabular}{l r r r @{\hspace{1.6em}} l r r r}
        \toprule
        \multicolumn{4}{c}{Opacity pruning} & \multicolumn{4}{c}{SH quantization (whole file)} \\
        \cmidrule(lr){1-4}\cmidrule(lr){5-8}
         & WM-PSNR & $r$ & $r{\geq}\tau$ & & WM-PSNR & $r$ & $r{\geq}\tau$ \\
        \midrule
        0\,\%  & 38.32 & 0.998 & 150 & FP32/16 & 38.32 & 0.998 & 150 \\
        10\,\% & 23.38 & 0.965 & 150 & INT12   & 38.25 & 0.998 & 150 \\
        30\,\% & 18.21 & 0.892 & 150 & INT10   & 37.35 & 0.998 & 150 \\
        50\,\% & 15.38 & 0.790 & 150 & INT8    & 32.08 & 0.994 & 150 \\
        80\,\% & 12.70 & 0.544 & 52  &         &       &       &     \\
        \bottomrule
    \end{tabular}
\end{table}

\subsection{Robustness to Post-processing}
\label{sec:robustness}

\paragraph{Non-adaptive post-processing.}
FP16 quantization is lossless for extraction, and after whole-file SH INT8 (every Gaussian's SH tensors, per-tensor
min--max, geometry untouched) all $150$ models verify with mean $r=0.994$ (Table~\ref{tab:robustness},
Figure~\ref{fig:qualitative}); quantizing the centres as well is a different attack, and the float16 cast above shows that SHield does not
survive it. Removing the lowest-opacity
Gaussians degrades WM-PSNR sharply but $r$ gradually: all models pass $\tau$ at $50\%$ pruning and $52$ of $150$ at
$80\%$, while costing the adversary at most $1.7$~dB at $30\%$ and $9.6$--$18.0$~dB at $80\%$. Near-threshold outcomes are seed-sensitive.

A white-box adversary can also target parameter bands (Table~\ref{tab:truncation}, Appendix~\ref{app:truncation}). For wm1, truncation-aware
extraction passes on all $30$ scenes at degree~$2$ but fails at degree~$1$ and~$0$, making truncation to degree~$\leq 1$ an effective and sometimes cheap
removal attack, costing $1.36$~dB on average at degree~$1$. Over all payloads $114$ of $150$ models fail at degree~$1$,
and none of the $36$ passes establishes image recovery: $30$ are wm2 models lifted by that payload's high null floor and $6$ are
wm3 models carrying two tiles, both consequences of the tiling index map. The coprime
stride keeps bit payloads decodable -- every bit decodes at every degree across all $42$ runs (margin $0.062$ at
degree~$0$) -- whereas a noise-free simulation of plain tiling gives $52$--$81\%$, near chance at degree~$0$.

\begin{table}[t]
    \centering
    \small
    \caption{\textbf{Adaptive fine-tuning attack.} $\Delta$PSNR: quality on \emph{held-out} views against the
    distributed model, i.e.\ the adversary's cost (training-view quality rises throughout). $r$ after
    re-association and at the key's stored indices (\emph{oracle}); passes at $\tau=0.6$.}
    \label{tab:attack_finetune}
    \setlength{\tabcolsep}{4pt}
    \begin{tabular}{l r r r r r r r r}
        \toprule
        & & & \multicolumn{2}{c}{$r$ re-associated} & \multicolumn{2}{c}{$r$ oracle indices} & & \\
        \cmidrule(lr){4-5}\cmidrule(lr){6-7}
        Fine-tuned & steps & $\Delta$PSNR & mean & min & mean & min & matched (\%) & passes \\
        \midrule
        SH only & 100 & $+$0.01 & 0.9615 & 0.9282 & 0.9615 & 0.9282 & 100.0 & 5/5 \\
        SH only & 500 & $-$0.04 & 0.8994 & 0.8458 & 0.8994 & 0.8458 & 100.0 & 5/5 \\
        SH only & 2,000 & $-$0.18 & 0.7740 & 0.6634 & 0.7740 & 0.6634 & 100.0 & 5/5 \\
        SH only & 5,000 & $-$0.39 & 0.5257 & 0.4312 & 0.5257 & 0.4312 & 100.0 & 1/2 \\
        SH only & 10,000 & $-$0.63 & 0.3946 & 0.3109 & 0.3946 & 0.3109 & 100.0 & 0/2 \\
        SH only & 20,000 & $-$0.95 & 0.3034 & 0.2367 & 0.3034 & 0.2367 & 100.0 & 0/2 \\
        SH + geometry & 2,000 & $-$1.00 & 0.4857 & 0.2971 & 0.8449 & 0.7428 & 12.6 & 1/5 \\
        \bottomrule
    \end{tabular}
\end{table}

\paragraph{Adaptive attacks.}
\label{sec:attack}
An adversary with the file and comparable training data can re-optimize the model under the ordinary 4DGS
photometric loss, with no key and no watermark term. We run this on five scenes for $100$--$20{,}000$ steps, over the
SH alone or over SH and geometry, and extract both after re-association and at the key's stored indices, which
separates an erased payload from an unmatched one. Both variants defeat verification (Table~\ref{tab:attack_finetune}).
Fine-tuning the SH erases the payload: every entry still matches, but mean $r$ falls from $0.774$ at $2{,}000$ steps
to $0.395$ at $10{,}000$. Fine-tuning geometry as well defeats verification at $2{,}000$ steps on four of five scenes
\emph{without} erasing the mark ($r=0.845$ at the stored indices): the marked centres move $0.32$--$0.66\,s_\text{nn}$,
far beyond $\epsilon$, and the fraction still inside $\epsilon$ predicts the match rate to within $0.0002$ on every
run. The attack costs $0.39$~dB on held-out views at the first SH-only budget that breaks verification and $1.00$~dB
with geometry, a price many would pay, so we do not claim robustness against adaptive fine-tuning; the geometry
variant shows that a defence needs position-robust matching, not a stronger payload.

\emph{Attacks on the matcher.} Three cheaper attacks target re-association rather than the payload
(Appendix~\ref{app:attack}). Duplicating Gaussians in place makes key entries ambiguous, and duplicating all of them
defeats the shipped matcher ($r=0.0285$, $0/30$) at the cost of doubling the file; taking the lowest-numbered
bit-identical candidate instead restores $r=0.9997$ at every duplication rate, and requiring a lone candidate to be bit-identical removes all
tolerance-induced false matches. A global translation or rotation matches nothing, but aligning the file to the
original model first recovers translation on all $30$ scenes and rotation on $25$. We report the shipped matcher
throughout so that results stay comparable.

\begin{table}
    \centering
    \caption{\textbf{Cross-threat-model reference with Table~1 of Mark4D~\citep{lee2026mark4d}.} Baselines are quoted;
    SHield is measured on our base models with the same message lengths but a different D-NeRF protocol.
    \textbf{Bold} marks the best value in each column; the row groups answer different threat models, so this is a
    cross-threat-model reference rather than a ranking. Acc.: bit accuracy (\%); ``--'': unsupported. Baseline decoders:
   \citep{ma2022cin,zhu2018hidden,tancik2020stegastamp,fernandez2022ssl,zhang2019rivagan,fernandez2024videoseal}.}
\label{tab:mark4d}
    \footnotesize
    \setlength{\tabcolsep}{2.0pt}
    \resizebox{0.93\textwidth}{!}{% about 0.89 scale: text stays near 7.2 pt
    \begin{tabular}{@{}l cccc cccc cccc@{}}
        \toprule
        & \multicolumn{4}{c}{32 bits} & \multicolumn{4}{c}{48 bits} & \multicolumn{4}{c}{64 bits} \\
        \cmidrule(lr){2-5}\cmidrule(lr){6-9}\cmidrule(lr){10-13}
        Method & Acc. & PSNR & SSIM & LPIPS & Acc. & PSNR & SSIM & LPIPS & Acc. & PSNR & SSIM & LPIPS \\
        \midrule
        \multicolumn{13}{@{}l}{\textit{Frame-level, render domain: decode rendered frames}} \\
        CIN         & 67.66 & 35.62 & .9814 & .0261 & 64.22 & 34.96 & .9794 & .0290 & 59.64 & 34.49 & .9780 & .0278 \\
        HiDDeN      & 82.21 & 31.48 & .9550 & .0425 & 79.11 & 31.07 & .9566 & .0375 & 76.85 & 31.67 & .9556 & .0478 \\
        StegaStamp  & 85.12 & 33.03 & .9793 & .0472 & 80.59 & 32.64 & .9718 & .0554 & 76.09 & 32.32 & .9707 & .0540 \\
        SSL         & 85.24 & 30.73 & .9651 & .0436 & 81.52 & 29.98 & .9620 & .0486 & 78.37 & 29.56 & .9596 & .0504 \\
        GaussianMarker     & 86.36 & 32.94 & .9627 & .0405 & 82.44 & 32.17 & .9597 & .0459 & 80.92 & 33.59 & .9688 & .0440 \\
        GuardSplat         & 88.33 & 38.69 & .9934 & .0082 & 84.74 & 37.48 & .9921 & .0105 & 81.30 & 37.49 & .9916 & .0118 \\
        3D-GSW             & 90.46 & 31.98 & .9651 & .0321 & 87.49 & 32.09 & .9604 & .0411 & 83.36 & 30.87 & .9504 & .0512 \\
Mark4D & 96.34 & 42.32 & \textbf{.9960} & \textbf{.0018} & 94.54 & 41.45 & \textbf{.9954} & \textbf{.0023} & 92.71 & 41.27 & .9954 & \textbf{.0022} \\
        \midrule
        \multicolumn{13}{@{}l}{\textit{Video-level, render domain: decode rendered clips}} \\
        RivaGAN     & 80.51 & 30.41 & .9556 & .0506 & -- & -- & -- & -- & -- & -- & -- & -- \\
        VideoSeal   & 88.21 & 36.29 & .9811 & .0103 & 82.72 & 35.55 & .9780 & .0122 & 77.98 & 36.16 & .9810 & .0211 \\
Mark4D & 96.96 & 42.32 & \textbf{.9960} & \textbf{.0018} & 95.02 & 41.45 & \textbf{.9954} & \textbf{.0023} & 92.77 & 41.27 & .9954 & \textbf{.0022} \\
        \midrule
        \multicolumn{13}{@{}l}{\textit{Model file, non-blind: read the file with the owner's key}} \\
SHield (ours) & \textbf{100.00} & \textbf{44.40} & .9958 & .0036 & \textbf{100.00} & \textbf{44.16} & \textbf{.9954} & .0039 & \textbf{100.00} & \textbf{44.32} & \textbf{.9958} & .0036 \\
        \bottomrule
    \end{tabular}}
\end{table}

\subsection{Reference Comparison with Mark4D}
\label{sec:baseline_comparison}
Table~\ref{tab:mark4d} places SHield beside \citet{lee2026mark4d} on the same scenes and message lengths;
baselines are quoted and the underlying D-NeRF protocol differs (Appendix~\ref{app:mark4d}). SHield decodes every bit in all $42$ runs at
$\alpha_0{=}0.1$, rendering at $44.16$--$44.40$~dB against the original model. SHield holds the best bit accuracy and PSNR in
Table~\ref{tab:mark4d} and Mark4D the best SSIM and LPIPS, but the accuracy in particular follows from the access
model rather than from a better watermark: closed-form subtraction of the stored coefficients on an unmodified file
recovers the symbols exactly, so anything below $100\%$ would indicate a bug. The comparison is a cross-threat-model
reference, not a ranking, and it measures fidelity and capacity rather than robustness; under a strict
instantiation of Mark4D's own model-level distortion (SH noise, $20\%$ of Gaussians removed at random), SHield
still decodes all $32$ bits on all eight D-NeRF scenes.

% ─────────────────────────────────────────────────────────────────────────────
\section{Limitations}
\label{sec:limitations}
SHield protects the model file; content that circulates only as rendered video calls for a render-domain
watermark, and the two are complementary. Verification is non-blind: the owner keeps a key of $0.5$--$18.1$~MB
(median $2.8$~MB) per model and registers it, with the payload, before distribution. Re-association expects
essentially unchanged centres; rigid transforms can be undone by alignment, but noisy or half-precision centres
cannot, so position-robust matching, which would also blunt the geometry fine-tuning attack, is the most useful
next step.

The uncentred statistic has a payload-dependent floor: on an unmarked model the recovery collapses to the payload's
per-channel mean colour, which correlates with the payload at $\sqrt{\eta^2}$ ($0.029$--$0.368$ for our payloads,
undefined for greyscale wm3). Payloads therefore need $\eta^2$ well below $\tau^2=0.36$, checked once at
registration, or the centred statistic. Our robustness claims cover non-adaptive post-processing (for SH
truncation, bit payloads only); we report the adaptive and truncation attacks on image payloads without claiming robustness to them. Dedicated 3DGS compression pipelines~\citep{in2025compmarkgs,wang2026robust3dgsw}
remain untested, temporal stability is measured with a rendering-cost proxy rather than a video benchmark, and
the anchored optimization uses the training images, although anchoring to the owner's own renders works about as
well.

\section{Conclusion}
\label{sec:conclusion}
SHield watermarks redistributed 4DGS assets without a decoder: a key-selected, tiled payload in the canonical SH
coefficients is time- and view-invariant, and extraction is closed-form arithmetic on the model file. It resists INT8
quantization and opacity pruning, while the evaluation maps where it stops -- unchanged centres are required, SH
truncation removes image payloads, and an adversary who re-trains the model can remove the mark at a small quality cost. Those limits
name the next steps: position-robust re-association, statistics without payload-dependent null floors, and pairing
file-level marks with render-domain ones, which protect a different surface.

\subsection*{AI Use Statement}

% (This section is \textbf{required} and does not count toward the page limit.)

In this work, we used large language model (LLM) assistants to help with writing and editing portions of the paper text. All AI-assisted text has been reviewed, verified, and revised by the authors. All experimental results, claims, and conclusions are the sole responsibility of the authors. We take full responsibility for the final content of this work.

\subsection*{Ethics Statement}

% (This section is \textbf{recommended} and does not count toward the page limit.)

This work proposes a watermarking method for 4D Gaussian Splatting models intended to help content creators establish ownership of their assets. It does not involve human subjects, personally identifiable information, or sensitive datasets. As with other watermarking methods, the same mechanism could be used to hide unrelated data in shared models, and a description of the carrier could inform removal attempts; we report the attacks that defeat the image payload (\S~\ref{sec:robustness}) so that users can judge its limits.

\subsection*{Reproducibility Statement}

% (This section is \textbf{recommended} and does not count toward the page limit.)

The experimental setup, hyperparameters and dataset configurations are described in \S~\ref{sec:experiments} and Appendix~\ref{app:impl}. The Mark4D comparison setup, including its differences from the Mark4D protocol, is detailed in Appendix~\ref{app:mark4d}, the re-association validation in Appendix~\ref{app:reassoc}, and the verification protocol, including the in-sample choice of $\tau$, the scene-disjoint split and its seed, and the fresh trials, in Appendix~\ref{app:verification}. All experiments use the publicly available D-NeRF~\citep{pumarola2021d}, HyperNeRF~\citep{park2021hypernerf} and DyNeRF~\citep{li2022neural} datasets. Per-scene, per-payload and per-seed breakdowns are provided in the appendix.

% ─────────────────────────────────────────────────────────────────────────────
\bibliography{references}

@inproceedings{wu20244d,
  title     = {{4D} {G}aussian {S}platting for Real-Time Dynamic Scene Rendering},
  author    = {Wu, Guanjun and Yi, Taoran and Fang, Jiemin and Xie, Lingxi and
               Zhang, Xiaopeng and Wei, Wei and Liu, Wenyu and Tian, Qi and Wang, Xinggang},
  booktitle = {Proceedings of the IEEE/CVF Conference on Computer Vision and
               Pattern Recognition},
  pages     = {20310--20320},
  year      = {2024},
}

@inproceedings{yang2024real,
  title     = {Real-time Photorealistic Dynamic Scene Representation and Rendering
               with {4D} {G}aussian {S}platting},
  author    = {Yang, Zeyu and Yang, Hongye and Pan, Zijie and Zhu, Xiatian and Zhang, Li},
  booktitle = {International Conference on Learning Representations},
  year      = {2024},
}

@article{kerbl20233d,
  title   = {{3D} {G}aussian {S}platting for Real-Time Radiance Field Rendering},
  author  = {Kerbl, Bernhard and Kopanas, Georgios and Leimk{\"u}hler, Thomas and
             Drettakis, George},
  journal = {ACM Transactions on Graphics},
  volume  = {42},
  number  = {4},
  pages   = {1--14},
  year    = {2023},
}

@article{park2021hypernerf,
  title   = {{HyperNeRF}: A Higher-Dimensional Representation for Topologically Varying Neural Radiance Fields},
  author  = {Park, Keunhong and Sinha, Utkarsh and Hedman, Peter and Barron, Jonathan T. and
             Bouaziz, Sofien and Goldman, Dan B and Martin-Brualla, Ricardo and Seitz, Steven M.},
  journal = {ACM Transactions on Graphics},
  volume  = {40},
  number  = {6},
  pages   = {1--14},
  year    = {2021},
}

@inproceedings{pumarola2021d,
  title     = {{D-NeRF}: Neural Radiance Fields for Dynamic Scenes},
  author    = {Pumarola, Albert and Corona, Enric and Pons-Moll, Gerard and
               Moreno-Noguer, Francesc},
  booktitle = {Proceedings of the IEEE/CVF Conference on Computer Vision and
               Pattern Recognition},
  pages     = {10318--10327},
  year      = {2021},
}

@inproceedings{lee2026mark4d,
  title     = {{Mark4D}: Temporally-Consistent Watermarking for {4D} {G}aussian {S}platting},
  author    = {Lee, Jaejin and Jeong, Minjae and Park, Joonhyuk and Hwang, Yechan and
               Baek, Seunghun and Kim, Won Hwa},
  booktitle = {Proceedings of the IEEE/CVF Conference on Computer Vision and
               Pattern Recognition},
  pages     = {39393--39402},
  year      = {2026},
}

@article{zhou20264dgsw,
  title   = {{4D-GSW}: Kinematic-Aware Spatio-Temporal Consistent Watermarking for
             {4D} {G}aussian {S}platting},
  author  = {Zhou, Sifan and Zhang, Hang and Wang, Yuhang and Li, Ming},
  journal = {arXiv preprint arXiv:2605.22342},
  year    = {2026},
}

@inproceedings{luo2023copyrnerf,
  title     = {{CopyRNeRF}: Protecting the {C}opyright of {N}eural {R}adiance {F}ields},
  author    = {Luo, Ziyuan and Guo, Qing and Cheung, Ka Chun and See, Simon and Wan, Renjie},
  booktitle = {Proceedings of the IEEE/CVF International Conference on Computer Vision},
  year      = {2023},
}

@inproceedings{li2023steganerf,
  title     = {{StegaNeRF}: Embedding Invisible Information within Neural Radiance Fields},
  author    = {Li, Chenxin and Feng, Brandon Y. and Fan, Zhiwen and Pan, Panwang and Wang, Zhangyang},
  booktitle = {Proceedings of the IEEE/CVF International Conference on Computer Vision},
  year      = {2023},
}

@inproceedings{zhu2018hidden,
  title     = {{HiDDeN}: Hiding Data With Deep Networks},
  author    = {Zhu, Jiren and Kaplan, Russell and Johnson, Justin and Fei-Fei, Li},
  booktitle = {European Conference on Computer Vision},
  pages     = {682--697},
  year      = {2018},
}

@inproceedings{tancik2020stegastamp,
  title     = {{StegaStamp}: Invisible Hyperlinks in Physical Photographs},
  author    = {Tancik, Matthew and Mildenhall, Ben and Ng, Ren},
  booktitle = {Proceedings of the IEEE/CVF Conference on Computer Vision and
               Pattern Recognition},
  pages     = {2117--2126},
  year      = {2020},
}

@inproceedings{zhang2018unreasonable,
  title     = {The Unreasonable Effectiveness of Deep Features as a Perceptual Metric},
  author    = {Zhang, Richard and Isola, Phillip and Efros, Alexei A. and
               Shechtman, Eli and Wang, Oliver},
  booktitle = {Proceedings of the IEEE/CVF Conference on Computer Vision and
               Pattern Recognition},
  pages     = {586--595},
  year      = {2018},
}

@article{wang2004image,
  title   = {Image Quality Assessment: From Error Visibility to Structural Similarity},
  author  = {Wang, Zhou and Bovik, Alan C. and Sheikh, Hamid R. and Simoncelli, Eero P.},
  journal = {IEEE Transactions on Image Processing},
  volume  = {13},
  number  = {4},
  pages   = {600--612},
  year    = {2004},
}

@article{cox2007digital,
  title   = {Digital Watermarking and Steganography},
  author  = {Cox, Ingemar and Miller, Matthew and Bloom, Jeffrey and Fridrich, Jessica
             and Kalker, Ton},
  journal = {Morgan Kaufmann Publishers},
  year    = {2007},
}

@book{peyre2019computational,
  title     = {Computational Optimal Transport},
  author    = {Peyr{\'e}, Gabriel and Cuturi, Marco},
  publisher = {Foundations and Trends in Machine Learning},
  year      = {2019},
}

@inproceedings{ren2026keyss,
  title     = {All That Glitters Is Not Gold: Key-Secured {3D} Secrets within {3D} Gaussian Splatting},
  author    = {Ren, Yan and Lu, Shilin and Kong, Adams Wai-Kin},
  booktitle = {International Conference on Learning Representations (ICLR)},
  year      = {2026},
}

@inproceedings{chen2025guardsplat,
  title     = {GuardSplat: Efficient and Robust Watermarking for {3D} Gaussian Splatting},
  author    = {Chen, Zixuan and Wang, Guangcong and Zhu, Jiahao and Lai, Jianhuang and Xie, Xiaohua},
  booktitle = {Proceedings of the IEEE/CVF Conference on Computer Vision and Pattern Recognition (CVPR)},
  year      = {2025}
}

@inproceedings{huang2024gaussianmarker,
  title     = {GaussianMarker: Uncertainty-Aware Copyright Protection of {3D} Gaussian Splatting},
  author    = {Huang, Xiufeng and Li, Ruiqi and Cheung, Yiu-ming and Cheung, Ka Chun and See, Simon and Wan, Renjie},
  booktitle = {Advances in Neural Information Processing Systems (NeurIPS)},
  year      = {2024}
}

@inproceedings{jang20253dgsw,
  title     = {{3D-GSW}: {3D} Gaussian Splatting for Robust Watermarking},
  author    = {Jang, Youngdong and Park, Hyunje and Yang, Feng and Ko, Heeju and Choo, Euijin and Kim, Sangpil},
  booktitle = {Proceedings of the IEEE/CVF Conference on Computer Vision and Pattern Recognition (CVPR)},
  pages     = {5938--5948},
  year      = {2025}
}

@inproceedings{ma2022cin,
  title     = {Towards Blind Watermarking: Combining Invertible and Non-invertible Mechanisms},
  author    = {Ma, Rui and Guo, Mengxi and Hou, Yi and Yang, Fan and Li, Yuan and Jia, Huizhu and Xie, Xiaodong},
  booktitle = {Proceedings of the ACM International Conference on Multimedia (MM)},
  year      = {2022}
}

@inproceedings{fernandez2022ssl,
  title     = {Watermarking Images in Self-Supervised Latent Spaces},
  author    = {Fernandez, Pierre and Sablayrolles, Alexandre and Furon, Teddy and J{\'e}gou, Herv{\'e} and Douze, Matthijs},
  booktitle = {IEEE International Conference on Acoustics, Speech and Signal Processing (ICASSP)},
  year      = {2022}
}

@article{zhang2019rivagan,
  title   = {Robust Invisible Video Watermarking with Attention},
  author  = {Zhang, Kevin Alex and Xu, Lei and Cuesta-Infante, Alfredo and Veeramachaneni, Kalyan},
  journal = {arXiv preprint arXiv:1909.01285},
  year    = {2019}
}

@article{fernandez2024videoseal,
  title   = {Video Seal: Open and Efficient Video Watermarking},
  author  = {Fernandez, Pierre and Elsahar, Hady and Yalniz, I. Zeki and Mourachko, Alexandre},
  journal = {arXiv preprint arXiv:2412.09492},
  year    = {2024}
}

@inproceedings{li2022neural,
  title     = {Neural 3D Video Synthesis from Multi-view Video},
  author    = {Li, Tianye and Slavcheva, Mira and Zollhoefer, Michael and Green, Simon and Lassner, Christoph and Kim, Changil and Schmidt, Tanner and Lovegrove, Steven and Goesele, Michael and Newcombe, Richard and Lv, Zhaoyang},
  booktitle = {Proceedings of the IEEE/CVF Conference on Computer Vision and Pattern Recognition (CVPR)},
  pages     = {5521--5531},
  year      = {2022}
}

@inproceedings{liu2025hideinmotion,
  title     = {Hide-in-Motion: Embedding Steganographic Copyright Information into {4D} {G}aussian Splatting Assets},
  author    = {Liu, Hengyu and Li, Chenxin and Pan, Wentao and Yang, Zhiqin and Yang, Yifeng and Liu, Yifan and Li, Wuyang and Yuan, Yixuan},
  booktitle = {IEEE International Conference on Robotics and Automation (ICRA)},
  pages     = {2694--2700},
  year      = {2025},
  doi       = {10.1109/ICRA55743.2025.11128285}
}

@inproceedings{zhang2024gshider,
  title     = {{GS-Hider}: Hiding Messages into {3D} {G}aussian Splatting},
  author    = {Zhang, Xuanyu and Meng, Jiarui and Li, Runyi and Xu, Zhipei and Zhang, Yongbing and Zhang, Jian},
  booktitle = {Advances in Neural Information Processing Systems (NeurIPS)},
  volume    = {37},
  year      = {2024}
}

@article{guo2026splats,
  title     = {Splats in Splats: Robust and Effective {3D} Steganography towards {G}aussian Splatting},
  author    = {Guo, Yijia and Huang, Wenkai and Li, Yang and Li, Gaolei and Zhang, Hang and Hu, Liwen and Li, Jianhua and Huang, Tiejun and Ma, Lei},
  journal   = {Proceedings of the AAAI Conference on Artificial Intelligence},
  volume    = {40},
  number    = {6},
  pages     = {4485--4493},
  year      = {2026}
}

@inproceedings{bae2024ed3dgs,
  title     = {Per-{G}aussian Embedding-Based Deformation for Deformable {3D} {G}aussian Splatting},
  author    = {Bae, Jeongmin and Kim, Seoha and Yun, Youngsik and Lee, Hahyun and Bang, Gun and Uh, Youngjung},
  booktitle = {European Conference on Computer Vision (ECCV)},
  pages     = {321--335},
  year      = {2024}
}

@inproceedings{meng2026whispersplat,
  title     = {{WhisperSplat}: Lossless Steganography in {3D} {G}aussian Splatting},
  author    = {Meng, Nicole and Sahu, Ronak and Yin, Miao and Shezan, Faysal Hossain and Lao, Yingjie},
  booktitle = {Proceedings of the 43rd International Conference on Machine Learning},
  series    = {PMLR},
  volume    = {306},
  year      = {2026}
}

@inproceedings{shi2026guard4d,
  title     = {{Guard4D}: Robust Watermarking for {4D} {G}aussian Splatting via Decoupled Decoding},
  author    = {Shi, Yingying and Zhou, Zhong and Zhou, Bin},
  booktitle = {Proceedings of the 35th International Joint Conference on Artificial Intelligence (IJCAI-ECAI)},
  year      = {2026}
}

@inproceedings{wang2026robust3dgsw,
  title     = {{Robust3DGSW}: Toward Robust Watermarking for Quantization-Aware {3D} {G}aussian Splatting},
  author    = {Wang, Boyu and Xia, Jun and Chen, Mingsong},
  booktitle = {Proceedings of the IEEE/CVF Conference on Computer Vision and Pattern Recognition (CVPR)},
  year      = {2026}
}

@article{in2025compmarkgs,
  title   = {{CompMarkGS}: Robust Watermarking for Compressed {3D} {G}aussian Splatting},
  author  = {In, Sumin and Jang, Youngdong and Jeong, Utae and Jang, MinHyuk and Park, Hyeongcheol and Park, Eunbyung and Kim, Sangpil},
  journal = {arXiv preprint arXiv:2503.12836},
  year    = {2025}
}

@article{guo2026splatspp,
  title   = {Splats in Splats++: Robust and Generalizable {3D} {G}aussian Splatting Steganography},
  author  = {Guo, Yijia and Huang, Wenkai and Hu, Tong and Li, Gaolei and Li, Yang and Hong, Yuxin and Hu, Liwen and Ling, Xitong and Li, Jianhua and Chen, Shengbo and Huang, Tiejun and Ma, Lei},
  journal = {arXiv preprint arXiv:2604.15862},
  year    = {2026}
}

@article{Kerckhoffs1883,
  author = {Kerckhoffs, A.},
  journal = {Journal des Sciences Militaires},
  pages = {161-191},
  title = {La cryptographie militaire},
  year = 1883
}
\bibliographystyle{iclr2027_conference}

% ─────────────────────────────────────────────────────────────────────────────
\appendix

\section{Implementation Details}
\label{app:impl}

\begin{algorithm}[h]
\footnotesize
\caption{SHield embedding}
\label{alg:embed}
\begin{algorithmic}[1]
\Require 4DGS model $\mathcal{G}$ with SH $\mathbf{h}$; payload $\mathbf{W}$; seed $\kappa$; scale $\alpha_0$; training views $\mathcal{V}$
\Ensure watermarked model $\tilde{\mathcal{G}}$, key $\mathcal{K}$
\State $K \gets$ adaptive count (Eq.~\ref{eq:kselect})
\State $\mathcal{S} \gets \textsc{Sample}(N, K; \kappa)$ \Comment{sorted, no replacement}
\State $n_\text{slots} \gets K\,C_\text{tot}$;\; resize $\mathbf{W}$ if $n_\text{wm}\!>\!n_\text{slots}$; $n_t \gets \lfloor n_\text{slots}/n_\text{wm}\rfloor$
\For{$c \in \{R,G,B\}$}
  \State $\mathbf{z}_c \gets \mathbf{W}_c - \overline{\mathbf{W}_c}$ \Comment{store $\overline{\mathbf{W}_c}$ in $\mathcal{K}$}
  \For{$s = 0 \ldots n_\text{slots}-1$}
    \State $t \gets \lfloor s/n_\text{wm}\rfloor$;\;\;
           $j \gets \bigl((s \bmod n_\text{wm}) + t\sigma\bigr) \bmod n_\text{wm}$
      \Comment{$\sigma{=}0$ for images}
        \State $\mathbf{E}_c\!\left[\lfloor s/C_\text{tot}\rfloor,\, s \bmod C_\text{tot}\right] \gets \mathbf{z}_c[j]$
  \EndFor
\EndFor
\State $\Delta \gets \alpha_0 \mathbf{E}$ \Comment{warm start}
\For{$i = 1 \ldots T$}
  \State $\mathbf{h}' \gets \textsc{Detach}(\mathbf{h})$;\;
         $\mathbf{h}'[\mathcal{S}] \gets \mathbf{h}[\mathcal{S}] + \Delta$
  \State render $n_v$ views of $\mathcal{V}$ with $\mathbf{h}'$ \Comment{grads to $\Delta$}
  \State $\Delta \gets \textsc{Adam}\big(\Delta, \nabla_\Delta[\mathcal{L}_\text{render} + \lambda_\text{wm}\mathcal{L}_\text{wm}]\big)$
      \Comment{Eq.~\ref{eq:loss}}
\EndFor
\State $\mathbf{h}[\mathcal{S}] \gets \mathbf{h}[\mathcal{S}] + \Delta$ \Comment{$\mathbf{a}$: opacity, scale, rotation}
\State $\mathcal{K} \gets (\mathcal{S}, \mathbf{h}_\text{orig}[\mathcal{S}], \boldsymbol{\mu}[\mathcal{S}], \mathbf{a}[\mathcal{S}], \alpha_0, \sigma, \overline{\mathbf{W}}, \mathbf{W})$
\end{algorithmic}
\end{algorithm}

\paragraph{SH Tensor Layout.}
In the official 4DGS implementation~\citep{wu20244d}, the SH appearance coefficients for each Gaussian are maintained as two \texttt{nn.Parameter} tensors: \texttt{\_features\_dc} $\in \mathbb{R}^{N \times 1 \times 3}$ (representing the degree-0 DC component) and \texttt{\_features\_rest} $\in \mathbb{R}^{N \times (C_\text{tot}-1) \times 3}$ (representing degrees 1 to $d_\text{max}$). 
During rendering, these tensors are concatenated to form a single $[N, C_\text{tot}, 3]$ tensor. Because SHield embeds the payload independently into each colour channel, the total available slot count per channel is $K C_\text{tot}$.

\paragraph{Key Format and Storage.}
The verification key file (\texttt{watermark\_info.npz}) is designed to be self-contained. It strictly stores:
\begin{itemize}
    \item[(i)] The selected Gaussian indices $\mathcal{S}$.
    \item[(ii)] The original, unmodified SH coefficients of the selected Gaussians.
    \item[(iii)] The canonical centres of the selected Gaussians, alongside the model's median nearest-neighbour spacing. 
    To resolve tie-breaks during re-association, we also store their original opacity, scale, and rotation.
    \item[(iv)] Payload metadata: shape, channel count, and per-channel means. 
    For bit messages, this includes the message itself, the coprime stride $\sigma$, and the whitening sequence.
    \item[(v)] Hyperparameters: the perturbation scale $\alpha_0$, the effective selection ratio, and the random seed.
\end{itemize}
Because the stored reference coefficients dominate the file footprint, the key size scales linearly with $K$ and remains completely independent of the training image resolution or video sequence length.
Across all 192 models evaluated in this paper, the compressed key sizes range from 0.514 to 18.116 MB (median 2.849 MB). 
Broken down by payload type, the median size is 4.175 MB for the 150 image-payload keys, and 1.756 MB for the 42 bit-payload keys. 
The identity fields stored for tie-breaking (iii) add approximately 24.3\% to the median key size.

\paragraph{SHield Settings.}
All experiments utilize the following fixed hyperparameter configuration unless otherwise specified:
\begin{center}
    \small
    \begin{tabular}{ll}
        \toprule
        \textbf{Hyperparameter} & \textbf{Value} \\
        \midrule
        Iterations $T$ & 500 \\
        Adam learning rate & $10^{-3}$ \\
        Loss weights ($\lambda_\text{wm}$, $\lambda_\text{ssim}$) & 100, 0.2 \\
        Views per step $n_v$ (pool per model) & 4 (32) \\
        Perturbation scale $\alpha_0$ & 0.1 \\
        Selection bounds ($\rho_\text{min}$, $\rho_\text{max}$); Min tiles $n_t^{\min}$ & 0.10, 0.50; 2 \\
        Key seed (selection, views; bits: message, whitening) & 42 \\
        Re-association tolerance $\epsilon$ & $0.01\,s_\text{nn}$ \\
        \bottomrule
    \end{tabular}
\end{center}

\paragraph{Anchored Slots.}
In Eq.~\ref{eq:lwm}, $\text{tile}(\mathbf{z}_c, n_t)$ repeats $\mathbf{z}_c$ exactly $n_t$ times, which is shorter than the $K C_\text{tot}$ slots of $\mathbf{e}_c$ whenever $n_\text{wm}$ does not divide $K C_\text{tot}$. 
For image payloads the loss therefore anchors the first $n_t n_\text{wm}$ slots, the ones fold-and-average reads (Eq.~\ref{eq:foldavg}); the trailing partial tile is written but never read, and is left to $\mathcal{L}_\text{render}$ alone. 
For bit payloads the target is expanded over all $K C_\text{tot}$ slots instead, because the coprime stride makes it more than a plain repeat of the symbol vector.

\paragraph{Runtime Overhead.}
\begin{itemize}
    \item \textbf{Tiled Embedding:} $<1$\,s (requires tensor arithmetic only).
    \item \textbf{Optimization (SHield):} $37$--$179$\,s (measured across the 12 logged seed-study re-embeddings detailed in Appendix~\ref{app:seeds}). 
    This assumes $500$ steps $\times$ $4$ views on a shared NVIDIA RTX 4090, inclusive of loading the model, training cameras, and ground-truth images.
    \item \textbf{Extraction:} $<1$\,s plus re-association time. 
    Extraction requires only a PLY file read, nearest-centre nearest-neighbor queries, and basic arithmetic, with zero rendering overhead.
\end{itemize}

\paragraph{Scenes and Evaluation Details.}
Our evaluation on the 17 HyperNeRF scenes spans three distinct capture splits: \textit{interp} (aleks-teapot, chickchicken, cut-lemon1, hand1-dense-v2, slice-banana, torchocolate); \textit{misc} (americano, cross-hands1, espresso, keyboard, oven-mitts, split-cookie, tamping); and \textit{vrig} (broom2, vrig-3dprinter, vrig-chicken, vrig-peel-banana).

For robustness evaluations:
\begin{itemize}
    \item \textbf{Quantization:} Whole-file INT$b$ quantization is applied as per-tensor min--max uniform quantization. 
    Specifically, the DC and higher-order SH tensors of the entire file are individually rounded to $2^b$ uniformly spaced levels between that tensor's minimum and maximum values. 
    FP16 is applied as a standard half-precision cast. 
    During extraction, the original, unquantized reference stored in the key is used.
    \item \textbf{Pruning Cost (\S~\ref{sec:robustness}):} The rendering cost of the adversary's pruning attack is measured on the wm1 models of \textit{bouncingballs}, \textit{hellwarrior}, \textit{hook}, \textit{jumpingjacks}, \textit{mutant}, \textit{standup}, \textit{trex} (D-NeRF), and \textit{coffee\_martini}, \textit{cook\_spinach}, \textit{flame\_steak} (DyNeRF). 
    We simulate physical deletion by forcing the opacity of the targeted Gaussians to zero, and then compare the PSNR against ground truth on 20 evenly spaced test views.
    \item \textbf{Rendering Residual (\S~\ref{sec:extraction}):} Residual evaluations utilize all 20 test views for D-NeRF, and 30 evenly spaced test views per scene for HyperNeRF and DyNeRF (yielding 830 total evaluated views).
\end{itemize}

% ─────────────────────────────────────────────────────────────────────────────
\section{Key Re-association}
\label{app:reassoc}

\paragraph{Re-association Algorithm.}
To securely re-associate the verification key with a suspect file, we employ a geometric matching strategy. 
For a given key entry $k$ with stored canonical centre $\boldsymbol{\mu}_k$, the algorithm proceeds as follows:
\begin{itemize}
    \item \textbf{Candidate Retrieval:} Identify all suspect Gaussians $j$ where $\|\tilde{\boldsymbol{\mu}}_j - \boldsymbol{\mu}_k\| \leq \epsilon$. 
    We strictly define $\epsilon = 0.01\,s_\text{nn}$, where $s_\text{nn}$ is the median nearest-neighbour spacing of the watermarked model (stored in the key).
    \item \textbf{Unique Match:} If exactly one candidate satisfies the distance threshold, the entry is assigned to it.
    \item \textbf{Tie-Breaking:} If multiple candidates fall within the $\epsilon$ tolerance, we consult static attributes (opacity, scale, and rotation). 
    The entry is assigned \textit{only} if exactly one candidate possesses attributes bit-identical to the key's stored values. 
    Otherwise, it is left unmatched.
    \item \textbf{Conflict Resolution:} If multiple key entries claim the same suspect Gaussian, only a single claimant with bit-identical attributes retains it; all others are left unmatched. 
    Note that static attributes are only used for tie-breaking and never override a unique geometric centre match.
    \item \textbf{Fallback:} An unmatched entry substitutes the stored reference coefficients in place of the suspect coefficients. 
    This yields a residual of zero in the extraction math, mathematically equivalent to a removed Gaussian.
\end{itemize}
This tolerance $\epsilon$ was fixed prior to the experiments described below and was not iteratively tuned.

\paragraph{Baseline Validation (Topological Robustness).}
Table~\ref{tab:reassoc} applies four conditions across all 192 models. 
Here, ``Exact'' denotes an extraction outcome perfectly identical to the index-based reference (e.g., the unmodified file for Conditions A and B, and in-place removal of the exact same Gaussians for C and D).

Without re-association, a simple row permutation completely destroys extraction (dropping image $r$ to between $-0.018$ and $0.075$, and bit accuracy to near-chance levels of $34$--$66\%$). 
With our algorithm applied to unmodified files, $99.66\%$ of key entries found exactly one candidate, while $0.34\%$ found several (up to $40$). 
Our attribute tie-breaker successfully resolved all $17{,}647$ ambiguous entries; notably, no model contained two Gaussians with perfectly identical centres and attributes. 
The rare false matches observed among removed entries were entirely tolerance-induced (i.e., a removed Gaussian happened to have a surviving neighbor within the $\epsilon$ radius).

\begin{table}[h]
    \centering
    \small
    \caption{\textbf{Geometric re-association performance across all 192 models} (150 image, 42 bit payloads). 
    Our centre-based matching algorithm demonstrates mathematically flawless structural recovery: across all adversarial conditions, surviving marked Gaussians were \emph{never} assigned to an incorrect row (0 mis-assignments). 
    \textbf{Metrics:} $\Delta$ denotes the maximum observed change in extraction fidelity (WM-PSNR and Pearson $r$) relative to theoretical, perfectly aligned in-place removal. 
    \textbf{Baseline Alignment:} Under Condition C (30\% opacity pruning), the geometrically recovered mean $r{=}0.892$ and WM-PSNR of $18.21$~dB perfectly reproduce the in-place baseline established in Table~\ref{tab:robustness}, proving our re-association module incurs virtually zero mathematical penalty.}
    \label{tab:reassoc}
    \setlength{\tabcolsep}{4pt}
    \resizebox{\textwidth}{!}{%
    \begin{tabular}{l r r r r r}
        \toprule
        Condition & exact & mis-assigned & removed entries falsely matched & $\Delta$ WM-PSNR / $r$ & bits \\
        \midrule
        A\; Unmodified file                         & 192/192 & 0 & -- & 0 / 0 & 100\% \\
        B\; Rows permuted                           & 192/192 & 0 & -- & 0 / 0 & 100\% \\
        C\; 30\% lowest opacity deleted, permuted   & 117/192 & 0 & 123 (0.008\%) & 0.006~dB / $1.5{\times}10^{-4}$ & 100\% \\
        D\; 20\% random deleted, permuted           & 28/192  & 0 & 1{,}189 (0.116\%) & 0.017~dB / $2.1{\times}10^{-4}$ & 100\% \\
        \bottomrule
    \end{tabular}}
\end{table}

\begin{figure}[h]
  \centering
  \includegraphics[width=\textwidth]{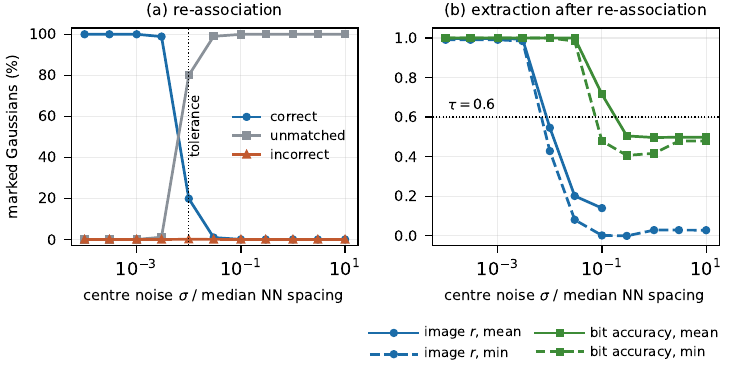}
  \caption{\textbf{Stress-testing geometric re-association under spatial noise.} 
  We evaluate the breakdown point of the matcher across all 192 models by injecting isotropic Gaussian noise into the canonical centres. 
  The fixed matching tolerance ($\epsilon{=}0.01\,s_\text{nn}$) is indicated by the dotted line. 
  \textbf{(a) Matching Dynamics:} The proportion of marked Gaussians that are correctly matched, safely left unmatched, or falsely matched to an incorrect Gaussian. 
  Crucially, the false-match rate remains strictly near zero across all noise levels. 
  \textbf{(b) Extraction Fidelity:} The resulting image-payload Pearson $r$ (mean and minimum over 150 models) and bit accuracy (mean and minimum over 42 runs) following the noisy re-association. 
  \textbf{Mathematical Bound:} The mean image $r$ becomes mathematically undefined beyond $\sigma{=}0.1\,s_\text{nn}$. 
  At this extreme noise level, every single entry is safely left unmatched, causing the greyscale payload's recovery to collapse into a flat mean-colour block with exactly zero variance.}
  \label{fig:e2}
\end{figure}

\paragraph{Stress-Testing Geometric Limits.}
To locate the exact breakdown point of the matcher, we applied isotropic Gaussian noise (per-axis s.d.\ $\sigma$) to every centre across all 192 models (with rows permuted, but SH and other attributes unchanged). 
Thresholds for material degradation were declared prior to scoring (Figure~\ref{fig:e2}).

All marked Gaussians are successfully matched up to $\sigma \leq 0.001\,s_\text{nn}$. 
At $0.003\,s_\text{nn}$ (roughly $0.8\%$ of the median Gaussian radius), $98.88\%$ of Gaussians remain matched. 
At $\sigma = \epsilon$, matching drops to $19.90\%$, causing image-payload verification to fail on 103 of the 150 models. 
(These measured rates follow theoretical $\chi_3$ predictions to within $0.03$ percentage points). 
Importantly, incorrect matches remain safely below $0.1\%$ of key entries at every noise level (peaking at $0.093\%$). 
Bit payloads prove more resilient: they only begin losing bits at $\sigma = 0.03\,s_\text{nn}$ (on a single model) and fall below $90\%$ mean accuracy at $0.1\,s_\text{nn}$.

Finally, casting the model centres to float16 leaves only $3.75\%$ of entries matched (none incorrectly). 
While this defeats image verification on 135 of 150 models, bit payloads still decode successfully. 
As a diagnostic test, relaxing the tolerance to $0.1\,s_\text{nn}$ matches $71.0\%$ of Gaussians after a float16 cast, but unacceptably increases false matches of removed entries to $1.79\%$ under $20\%$ deletion; no single tested tolerance securely handles both. 
When matching fails entirely, extraction defaults to the flat mean-colour block of an unmarked model. 
While this falls below $\tau$ for our tested payloads, it may not for payloads with high $\eta^2$ (Appendix~\ref{app:verification}). 
Therefore, we explicitly state that this re-association regime does not provide security against targeted, adaptive geometric manipulation.

% ─────────────────────────────────────────────────────────────────────────────
\section{Ownership Verification: Held-out Evaluation and Null Floor}
\label{app:verification}

\paragraph{Formalizing the Evaluation.}
To rigorously evaluate the watermark, we define two hypothesis states:
\begin{itemize}
    \item $H_1$ \textbf{(True Positive):} The correct key is applied to a watermarked model (either exactly as distributed or after post-processing).
    \item $H_0$ \textbf{(Null Hypotheses):} Verification is attempted without the correct mark. 
    We model this under three adversarial or coincidental conditions: \emph{other payload} (owner $i$'s key used on the same scene carrying owner $j$'s mark), \emph{wrong subset} (using the stored coefficients and payload with a randomly drawn selection of Gaussians, simulating a partially compromised key), and \emph{unmarked} (applying the owner's key to a clean, unmodified model).
\end{itemize}

\paragraph{Strict Evaluation Protocol.}
An initial in-sample analysis computed $1{,}800$ extractions across all $150$ models ($1{,}770$ yielded a defined $r$; the $30$ unmarked-model scores for the greyscale wm3 are mathematically undefined). 
From this, we selected a decision threshold of $\tau=0.6$.

To prevent data leakage, we subsequently split the $30$ scenes into $14$ calibration scenes ($70$ models) and $16$ evaluation scenes ($80$ models), stratified by dataset using a fixed seed ($20260916$). 
We locked $\tau=0.6$ unchanged. 
(The calibration scores naturally leave a wide margin around this threshold: max $H_0 = 0.4972$, min $H_1 = 0.8169$). 
The split, the decision rule ($r \geq \tau$, treating undefined $r$ as a non-detection), and the fresh-trial design were all fixed before any fresh trial was scored. 
We make no such claim for $\tau$ itself: it was chosen in-sample, as stated above, and only then carried unchanged onto scene-disjoint data.

The final evaluation consists of \emph{retrospective} scores (the original conditions on the evaluation scenes) and \emph{fresh} trials drawn exclusively for this analysis. 
The fresh trials include $400$ true $H_1$ trials ($5$ per model, under $30\%$ random Gaussian deletion, row permutation, and key re-association) and $1{,}600$ null $H_0$ trials ($20$ per model, drawing wrong subsets from new random seeds).

\begin{table}[h]
    \centering
    \small
    \caption{\textbf{Ownership verification on the held-out evaluation set} (16 scenes, 80 models, threshold $\tau{=}0.6$). 
    SHield achieves a perfect 100\% detection rate across all true positive ($H_1$) conditions while successfully registering zero false positives across all null ($H_0$) conditions, including 1,600 fresh adversarial trials. 
    \textbf{Decision Rule:} An undefined Pearson $r$ strictly defaults to a non-detection. 
    \textbf{Statistical Bounds \& Transparency:} Because multiple trials within a single model inherently share payloads and selections, they are statistically correlated. 
    Consequently, naive per-trial bounds (e.g., $0.19\%$ for the 1,600 fresh nulls) mathematically underestimate the true false-positive rate. 
    By strictly treating models or scenes as independent units, we derive nominal one-sided 95\% Clopper--Pearson upper bounds of $3.68\%$ per model ($n{=}80$) and $17.07\%$ per scene ($n{=}16$). 
    We explicitly report these boundaries, acknowledging their imperfect independence assumptions, to ensure absolute statistical rigor (\S~\ref{app:verification}).}
    \label{tab:verification}
    \setlength{\tabcolsep}{4pt}
    \resizebox{\textwidth}{!}{%
    \begin{tabular}{l l r r r r r}
        \toprule
         & Condition & $n$ & min & median & max & decisions at $\tau$ \\
        \midrule
        \multirow{3}{*}{\makecell[l]{$H_1$,\\retrospective}}
         & correct key, distributed model & 80 & 0.9910 & 0.9991 & 1.0000 & 80/80 detected \\
         & \quad after INT8               & 80 & 0.9766 & 0.9963 & 0.9995 & 80/80 detected \\
         & \quad after $30\%$ pruning     & 80 & 0.8147 & 0.9103 & 0.9737 & 80/80 detected \\
        \midrule
        \multirow{3}{*}{\makecell[l]{$H_0$,\\retrospective}}
         & other payload, same scene      & 320 & $-$0.0951 & 0.0060 & 0.3170 & 0 false positives \\
         & wrong subset                   & 320 & 0.0382 & 0.1417 & 0.4858 & 0 false positives \\
         & unmarked model (64 defined)    & 80  & 0.0285 & 0.1400 & 0.3681 & 0 false positives \\
        \midrule
        $H_1$, fresh & $30\%$ random deletion, permuted, re-associated & 400 & 0.8192 & 0.9056 & 0.9677 & 400/400 detected \\
        $H_0$, fresh & wrong subset, new seeds                         & 1{,}600 & 0.0410 & 0.1413 & 0.4981 & 0 false positives \\
        \bottomrule
    \end{tabular}}
\end{table}

\paragraph{Statistical Bounds and Assumptions.}
Our results demonstrate zero false positives across the $1{,}600$ fresh wrong-subset null trials from the $80$ evaluation models. 
While the exact per-trial Clopper--Pearson upper bound is nominally $0.19\%$, we emphasize that this assumes fully independent trials. 
Because the $20$ null trials for a given model share the same payload and selection, they are correlated; thus, this is not a strict bound on the per-trial false-positive rate.

Treating models or scenes as independent units, the nominal one-sided $95\%$ Clopper--Pearson upper bounds on producing any false positive are $3.68\%$ per model and $17.07\%$ per scene. 
Again, in the interest of complete statistical honesty, this independence assumption is imperfect because models share scenes and payloads (e.g., the five models of a scene share its base model). 
These figures represent our rigorous empirical test bounds rather than generalized deployment-level false-positive rates, especially given that the null distribution is heavily payload-dependent.

\paragraph{Sensitivity to the Threshold.}
Because $\tau$ was chosen in-sample, the held-out result could in principle depend on that choice. 
It does not, in a concrete sense. 
Across the whole evaluation set ($640$ $H_1$ and $2320$ $H_0$ trials, $16$ of the latter undefined), the largest null score is $0.4981$ and the smallest true score $0.8147$. 
Every decision reported above is therefore unchanged for any threshold in $(0.498, 0.815]$, a band $0.317$ wide; $\tau=0.6$ sits $32\%$ into it, with $0.102$ of headroom below and $0.215$ above. 
Repeating the evaluation with a threshold taken from the calibration scenes alone, the midpoint $0.6570$ of their own $H_0$/$H_1$ gap, which uses no evaluation data, accepts the same $640$ true trials and yields the same zero false positives. 
The in-sample choice of $\tau$ therefore does not carry the result: any threshold in a band roughly a third of the score range wide gives the identical outcome. 
This does not remove the payload-dependence of the null floor below, which is what bounds how wide that band is for a given payload.

\paragraph{The Payload-Dependent Null Floor.}
During extraction, SHield adds back the stored per-channel mean. 
If the file carries no watermark information (e.g., an unmarked model, or all entries remain unmatched due to geometric noise), the recovery mathematically collapses into a flat block of that mean colour. 
The Pearson correlation of this flat block against the original payload is exactly $\sqrt{\eta^2}$, where $\eta^2$ represents the fraction of the payload's pixel variance that lies between its channel means.

For our tested payloads, $\sqrt{\eta^2}$ is $0.029$ (wm1), $0.368$ (wm2), $0.214$ (wm4), and $0.067$ (wm5), which exactly matches the measured unmarked-model scores (wm3 is greyscale, rendering the floor undefined). 
This variance floor shapes the other nulls as well: all $318$ fresh wrong-subset trials that scored $r \geq 0.4$ utilized the wm2 payload. 
Crucially, a payload with $\eta^2 \geq \tau^2 = 0.36$ would be automatically falsely accepted from an unmarked model. 
This high floor also artificially inflates the scores of partially destroyed marks (e.g., wm2 crossing the threshold at SH degree 1; Table~\ref{tab:truncation}), which does not indicate successful visual recovery. 
(The $6$ wm3 passes at SH degree 1 are not a channel-mean effect: they are the six scenes whose wm3 embedding carries two tiles, where the tiling index map attenuates pixels rather than erasing them (Appendix~\ref{app:stride}). They likewise do not establish visual recovery.)

Therefore, our false-positive guarantees apply solely to the tested payloads and our specific protocol. 
For practical deployments, payloads must be explicitly screened to ensure $\eta^2$ is well below $\tau^2$. 
We deliberately do not alter the statistic in this paper to mask this property. 
Figure~\ref{fig:qualitative} illustrates extractions from the models of Figure~\ref{fig:diffmaps}, visually confirming the flat recovery from an unmarked model.

% ─────────────────────────────────────────────────────────────────────────────
\section{Carrier Selection}
\label{app:carrier}

\paragraph{Theoretical Advantage vs.\ Empirical Cost.}
As established in Section~\ref{sec:method}, embedding the watermark into canonical SH coefficients provides a powerful theoretical advantage: because these coefficients are queried at every timestamp, a single write covers the entire dynamic sequence. 
However, identifying where a watermark \emph{can} be mathematically placed does not dictate where it is visually \emph{cheapest} to place.

\begin{figure}[h]
  \centering
  \includegraphics[width=\textwidth]{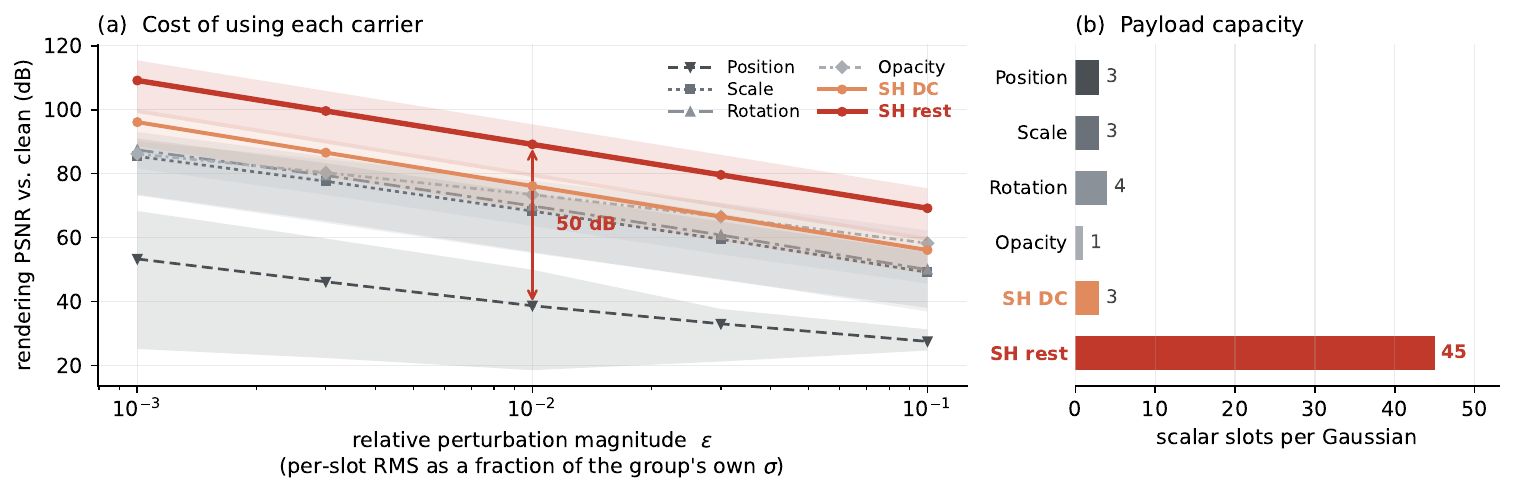}
  \caption{\textbf{Cost and capacity of Gaussian parameter groups as watermark carriers.} 
  This ablation systematically isolates the rendering penalty of embedding into different parameter spaces, decisively justifying our selection of higher-order SH bands. 
  \textbf{Ablation Setup:} The identical tiled, zero-mean payload is injected into each respective parameter group for the exact same key-selected subset ($K{=}10\%$). 
  To ensure a mathematically fair comparison across disparate parameter ranges, the perturbation is normalized such that its per-slot RMS strictly equals $\varepsilon$ times that specific group's natural standard deviation. 
  \textbf{(a) Empirical Rendering Cost:} PSNR measured against the clean render (where higher indicates a less visually destructive carrier), averaged over five scenes (four D-NeRF, one DyNeRF) with min--max variance shaded. 
  At $\varepsilon{=}0.01$, perturbing spatial positions triggers a catastrophic rendering penalty ($38.7$~dB). 
  In stark contrast, the higher-order SH bands achieve a near-lossless $89.2$~dB, massively outperforming the DC band ($76.1$~dB) and geometric attributes like scale, rotation, and opacity ($68$--$73$~dB). 
  \textbf{(b) Embedding Capacity:} The total available scalar slots per Gaussian, highlighting the immense capacity advantage ($45$ slots) of the higher-order SH bands.}
  \label{fig:carrier}
\end{figure}

\paragraph{Cost Isolation Experiment.}
Figure~\ref{fig:carrier} addresses this by systematically isolating the empirical rendering cost of each potential carrier. 
In this ablation, the payload signal and the selected subset of Gaussians are held strictly constant. 
We then individually apply the perturbation across different parameter groups (position, scale, rotation, opacity, and SH bands). 
To ensure a fair comparison that does not artificially penalize parameters with narrow natural ranges, each perturbation is normalized, scaled to the same relative multiple ($\varepsilon$) of that specific parameter group's own standard deviation.

\paragraph{Results and Temporal Stability.}
The empirical ordering of rendering costs remains perfectly consistent across every scene and every $\varepsilon$ value in our sweep. 
While the absolute PSNR values for most parameter groups remain well above the threshold of human perceptibility, the relative gaps between them are substantial and dictate our design choice.

Crucially, perturbing canonical positions incurs a massive PSNR penalty compared to perturbing higher-order SH bands. 
Beyond static image quality degradation, geometry introduces a severe temporal vulnerability in 4DGS: because the deformation network continuously propagates canonical positions to every timestamp, a static spatial perturbation would inadvertently cause the rendered content to warp or shift over time. 
Consequently, higher-order SH bands provide the optimal combination of high embedding capacity, minimal static rendering penalty, and reliable temporal stability.

% ─────────────────────────────────────────────────────────────────────────────
\section{Coprime Stride Interleaving}
\label{app:stride}

\begin{figure}[h]
  \centering
  \includegraphics[width=\textwidth]{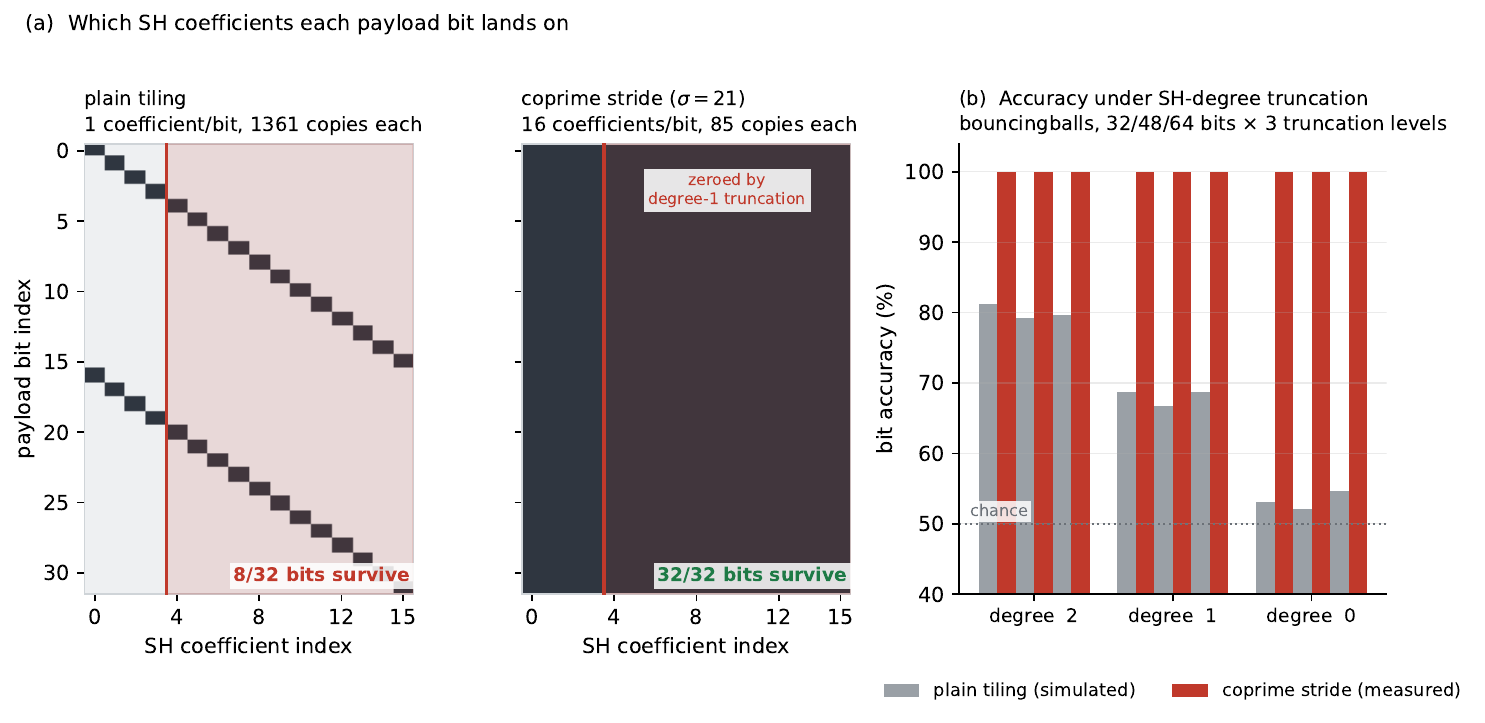}
  \caption{\textbf{Defeating targeted SH band erasure via coprime stride interleaving.} 
  This figure mathematically demonstrates how our coprime stride prevents the catastrophic erasure of bit payloads when adversaries systematically truncate higher-order SH bands. 
  In our flattened 1D formulation, slot $s$ inherently maps to SH coefficient $s \bmod C_\text{tot}$ (where $C_\text{tot}{=}16$). 
  \textbf{(a) Vulnerability vs.\ Defense:} We visualize a 32-bit message embedded into the \textit{bouncingballs} scene (dark slots denote at least one embedded copy). 
  \emph{Naive Tiling:} Because $C_\text{tot}{=}16$ perfectly divides all baseline message lengths ($L \in \{32, 48, 64\}$), plain tiling collapses all ${\approx}1{,}360$ redundant copies of a specific bit into a \emph{single} SH coefficient. 
  Consequently, truncating the model to SH degree~1 (red) deterministically erases every single copy of 24 of the 32 bits. 
  \emph{Coprime Stride:} By shifting each tile $t$ by $t\sigma$ (where $\sigma$ is strictly coprime to both $L$ and $C_\text{tot}$), SHield uniformly distributes the redundant copies across all 16 coefficients. 
  Under the identical degree~1 truncation attack, the stride merely sacrifices $75\%$ of the copies globally, safely preserving ${\approx}340$ redundant copies per bit to cleanly resolve the final sign decision. 
  \textbf{(b) Empirical Resilience:} Validating this mathematical guarantee on \textit{bouncingballs}: SHield maintains a flawless $100\%$ measured bit accuracy across all truncation levels and message lengths. 
  In stark contrast, a simulated noise-free naive tiling degrades proportionally with band erasure, plunging to near-chance recovery ($52$--$55\%$) when truncated down to the DC band (degree~0).}
  \label{fig:stride}
\end{figure}

\paragraph{The Structural Threat of SH Truncation.}
Unlike operations that introduce statistical noise (e.g., quantization), SH-degree truncation structurally removes entire coefficient indices simultaneously. 
Figure~\ref{fig:stride} illustrates how this deterministic erasure affects the embedded payload, and demonstrates how a single integer stored in the key, the coprime stride, mathematically guarantees that bit messages remain decodable.

\paragraph{The Vulnerability of Plain Tiling.}
For bit messages, standard payload lengths ($L \in \{32, 48, 64\}$) are evenly divisible by the total number of SH coefficients ($C_\text{tot}=16$). 
Under plain tiling, this divisibility causes every single copy of a specific bit to map to the exact same SH coefficient index. 
If an adversary truncates the model's SH degree and drops that specific index, the bit is permanently and entirely erased.

In our noise-free simulation of plain tiling, accuracy under truncation plummeted to near-chance levels ($52$--$81\%$) across all $42$ runs. 
Because this erasure is driven purely by structural arithmetic rather than signal degradation, increasing the embedding amplitude $\alpha_0$ provides absolutely no defense against it.

\paragraph{The Coprime Stride Defense.}
To counter this vulnerability, SHield shifts each tile by an integer $\sigma$ that is coprime to both $L$ and $C_\text{tot}$: $21$, $31$ and $41$ for $L=32$, $48$ and $64$. 
Such a $\sigma$ exists for every $L$, since $\sigma{=}1$ satisfies both conditions, and the search starts near $L/\varphi$ so that consecutive tiles land far apart. 
This interleaving completely breaks the index alignment, spreading the copies of every individual bit evenly across all $16$ SH coefficients.

When an adversary truncates the model to degree $d$, every bit safely retains a fraction $(d{+}1)^2/C_\text{tot}$ of its original copies. 
Consequently, the decoding accuracy remains at $100\%$ across all degrees, while the decoding margin shrinks predictably in exact proportion to the surviving copies ($0.561$, $0.249$, and $0.062$ at degrees $2$, $1$, and $0$, respectively; Table~\ref{tab:truncation}).

\paragraph{Impact on Image Payloads.}
Image payloads keep plain tiling ($\sigma=0$), so the copies of payload index $j$ occupy the SH coefficient indices $(j + t\,n_\text{wm}) \bmod C_\text{tot}$ for $t = 0,\dots,n_t{-}1$, and each pixel therefore reaches $\min\bigl(n_t,\; C_\text{tot}/\gcd(n_\text{wm}, C_\text{tot})\bigr)$ distinct indices. 
Which failure mode applies is thus fixed by $n_\text{wm}$ and $n_t$, not by signal strength. 
For wm1, $n_\text{wm} = 128 \times 394 = 50{,}432$ is an exact multiple of $C_\text{tot}=16$, so every copy of a pixel lands on the \emph{same} index, exactly as for bit payloads under plain tiling. 
Truncation to degree $d$ then does not attenuate all pixels alike: it erases a fraction $1-(d{+}1)^2/C_\text{tot}$ of the pixels outright, whose recovery falls back to the stored channel mean, and returns the remaining ones exactly. 
Writing $p=(d{+}1)^2/C_\text{tot}$ for the surviving fraction, the Pearson correlation of such a recovery against the payload is $\sqrt{p}$, giving $0.750$, $0.500$ and $0.250$ at degrees $2$, $1$ and $0$ and matching the measured $0.749$, $0.500$ and $0.251$ (Table~\ref{tab:truncation}). 
Uniform attenuation would instead leave $r$ unchanged, Pearson correlation being scale-invariant.

For the other four payloads $n_\text{wm}$ is not a multiple of $C_\text{tot}$ ($\gcd = 2$), so once $n_t \geq 2$ each pixel's copies reach several indices and truncation attenuates every pixel instead of erasing some, which holds $r$ well above $\sqrt{p}$; at $n_t = 1$ there is one copy per pixel and the erasure behaviour returns. 
Replaying this index arithmetic over the stored payloads reproduces the measured truncation-aware $r$ of every wm1 and wm2 model to within $0.005$, and accounts for all $36$ threshold passes at degree~$1$ (Table~\ref{tab:truncation}): the $30$ wm2 passes, which carry $n_t \geq 2$ on every scene and are lifted further by that payload's $0.368$ channel-mean floor, and the $6$ wm3 passes, which are exactly the six scenes whose wm3 embedding carries two tiles ($r \approx 0.705$, against $r \approx 0.500$ on the $24$ single-tile scenes). 
Neither group indicates visual recovery; both are structural artefacts of the index map and the channel-mean floor (\S~\ref{sec:limitations}).

% ─────────────────────────────────────────────────────────────────────────────
\section{SH Truncation: Details}
\label{app:truncation}

\begin{table}
    \centering
    \caption{\textbf{SH-degree truncation: attack cost and survival, by payload type} ($\alpha_0{=}0.1$).
    \emph{Image payload wm1} ($30$ scenes): $\Delta$PSNR against the untruncated render (mean [worst scene]); mean $r$
    and threshold passes with truncation-aware extraction (same bands zeroed in the reference) and naive extraction.
    \emph{All five payloads}: passes of $150$ models, aware (naive); $^{\dagger}$the $36$ passes at degree~1 are $30$ wm2 and $6$ wm3,
    both accounted for by the tiling index map and, for wm2, its high null floor (Appendix~\ref{app:stride});
    none establishes image recovery. \emph{Bit payloads}: $42$ runs, accuracy, decoding
    margin (1 = unattacked) and PSNR of the $14$ truncated $L{=}32$ models against the untruncated clean render.
    Per dataset: Appendix~\ref{app:truncation}.}
\label{tab:truncation}
    \setlength{\tabcolsep}{3.6pt}
    \resizebox{\textwidth}{!}{%
    \begin{tabular}{c r r r r r r r r}
        \toprule
        & \multicolumn{4}{c}{Image payload wm1 (30 scenes)} & \multicolumn{1}{c}{All image payloads}
          & \multicolumn{3}{c}{Bit payloads (42 runs)} \\
        \cmidrule(lr){2-5}\cmidrule(lr){6-6}\cmidrule(lr){7-9}
        SH degree & $\Delta$PSNR [worst] & $r$ aware & passes aware & passes naive & passes aware (naive)
          & bit accuracy & margin & PSNR vs.\ clean ($L{=}32$) \\
        \midrule
        3 & 0.00               & 0.9997 & 30/30 & 30/30 & 150 (150) & 100\% & 0.998 & 44.40 \\
        2 & $-$0.76 [$-$3.41] & 0.749  & 30/30 & 19/30 & 150 (51)  & 100\% & 0.561 & 38.56 \\
        1 & $-$1.36 [$-$4.54] & 0.500  & 0/30  & 0/30  & 36$^{\dagger}$ (0) & 100\% & 0.249 & 36.69 \\
        0 & $-$2.27 [$-$9.66] & 0.251  & 0/30  & 0/30  & 0 (0)     & 100\% & 0.062 & 33.75 \\
        \bottomrule
    \end{tabular}}
\end{table}

\paragraph{Simulation Methodology and DyNeRF Nuance.}
To evaluate the visual impact of SH truncation, we simulate the attack by limiting the 4DGS rasterizer to SH degree $d$ during rendering. 
For D-NeRF and HyperNeRF, this rasterizer limit is mathematically identical to zeroing the stored SH bands above degree $d$ directly in the file (yielding a maximum pixel difference of $0$). 
However, on DyNeRF, simply zeroing the stored bands is insufficient (yielding pixel differences up to $0.37$) because the deformation network dynamically applies time-dependent offsets to all $16$ coefficients.

Crucially, because SHield's extraction is strictly a file-level operation, it operates directly on the stored parameter bands, bypassing these render-time offsets entirely. 
As expected, evaluating extraction at degree $3$ reproduces the baseline untruncated evaluations exactly. 
Table~\ref{tab:trunc_detail} provides a comprehensive dataset-level breakdown of these truncation effects.

\begin{table}[h]
    \centering
    \small
    \caption{\textbf{Dataset-level breakdown of robustness against SH-degree truncation} ($\alpha_0{=}0.1$). 
    This detailed ablation isolates the extraction fidelity and rendering penalties across D-NeRF, HyperNeRF, and DyNeRF under progressive SH band erasure. 
    \textbf{Image Payloads (wm1):} We report the rendering penalty ($\Delta$PSNR) against ground truth relative to the untruncated render, alongside the Pearson $r$ and threshold passes for both truncation-aware and naive extraction. 
    \textbf{Artifact Isolation:} Crucially, when an unmarked clean model is truncated in the exact same manner, its rendering score falls within a negligible $0.07$~dB of the watermarked model. 
    This mathematically proves that the observed visual degradation is fundamentally caused by the loss of higher-order SH detail, rather than the adversarial exposure of watermarking artifacts. 
    \textbf{Bit Payloads:} We report the PSNR of the truncated $L{=}32$ models evaluated against the clean untruncated render (using Mark4D test views). 
    Empowered by the coprime stride, bit decoding accuracy remains a flawless $100\%$ on every single run across all datasets and all truncation degrees.}
    \label{tab:trunc_detail}
    \setlength{\tabcolsep}{3.5pt}
    \resizebox{\textwidth}{!}{%
    \begin{tabular}{l c r r r r}
        \toprule
        Dataset & degree & $\Delta$PSNR (dB) & $r$ aware / naive & passes aware / naive & bit models: PSNR (dB) \\
        \midrule
        D-NeRF    & 2 & $-$0.523 & 0.749 / 0.628 & 7/7 / 5/7   & 43.24 \\
                  & 1 & $-$1.142 & 0.500 / 0.302 & 0/7 / 0/7   & 40.89 \\
                  & 0 & $-$2.410 & 0.251 / 0.076 & 0/7 / 0/7   & 37.40 \\
        HyperNeRF & 2 & $-$0.243 & 0.749 / 0.619 & 17/17 / 10/17 & -- \\
                  & 1 & $-$0.785 & 0.500 / 0.300 & 0/17 / 0/17 & -- \\
                  & 0 & $-$1.420 & 0.251 / 0.079 & 0/17 / 0/17 & -- \\
        DyNeRF    & 2 & $-$2.476 & 0.749 / 0.590 & 6/6 / 4/6   & 32.33 \\
                  & 1 & $-$3.240 & 0.500 / 0.279 & 0/6 / 0/6   & 31.09 \\
                  & 0 & $-$4.535 & 0.251 / 0.071 & 0/6 / 0/6   & 28.88 \\
        \bottomrule
    \end{tabular}}
\end{table}

\paragraph{Adversarial Rendering Cost.}
From an adversarial perspective, degree-$1$ truncation is a highly accessible removal attack for image payloads. 
The rendering penalty incurred by the attacker is remarkably low, costing less than $0.25$, $0.5$, and $1.0$~dB of PSNR on $3$, $10$, and $17$ of the $30$ evaluated scenes, respectively.

\paragraph{Verifying Bit Payload Survival (Sanity Check).}
Because maintaining $100\%$ bit accuracy at degree $0$ might initially appear counterintuitive, and could theoretically arise from a decoding math artifact, we conducted a rigorous sanity check. 
We attempted to decode two runs from entirely \emph{clean} (unmarked) models using the exact same registered keys.

On these clean models, the decoding accuracy correctly collapsed to near-chance levels ($53.1\%$ and $45.3\%$) with negligible decoding margins ($0.018$ and $0.017$). 
By contrast, the watermarked models maintained $100\%$ accuracy with margins of $0.064$ and $0.062$. 
This verifies that the degree-$0$ survival is a genuine structural recovery rather than a decoder artifact, though the surviving margin drops to roughly $3.6\times$ that of an unmarked model. 
We note that this sanity check is designed strictly to validate survival mechanics and does not serve as a formal false-positive estimate for bit payloads.

% ─────────────────────────────────────────────────────────────────────────────
\section{Results per Scene and Payload}
\label{app:multi_wm}

\begin{table}[h]
    \centering
    \small
    \caption{\textbf{Comprehensive per-scene evaluation of SHield ($\alpha_0{=}0.1$).} 
    This granular breakdown demonstrates SHield's exceptional consistency, achieving near-perfect payload recovery (Pearson $r \geq 0.997$) across all individual scenes and datasets regardless of topological complexity. 
    \textbf{Aggregation Setup:} Each row reports the mean performance across all five distinct payloads; HyperNeRF rows further aggregate the scenes within each respective capture split. 
    \textbf{Metric Isolation:} Absolute metrics (PSNR, SSIM, and LPIPS) are measured against ground-truth images, primarily reflecting the natural reconstruction difficulty of each specific scene. 
    Crucially, $\Delta$PSNR isolates the pure mathematical rendering penalty incurred by the watermark, cleanly separating our defense's visual footprint from pre-existing baseline artifacts.}
    \label{tab:per_scene}
    \resizebox{\textwidth}{!}{%
    \begin{tabular}{l r r r r r r r}
        \toprule
        Scene / Split & PSNR$_\text{clean}$ & PSNR$_\text{wm}$ & $\Delta$PSNR & SSIM$_\text{wm}$ & LPIPS$_\text{wm}$
          & WM-PSNR & Pearson\,$r$ \\
        \midrule
        \multicolumn{8}{l}{\textit{D-NeRF} (7 synthetic monocular scenes)} \\
        Bouncingballs & 40.378 & 40.265 & $-$0.113 & 0.9940 & 0.0112 & 41.39 & 0.9994 \\
        Hellwarrior & 28.731 & 28.709 & $-$0.022 & 0.9720 & 0.0296 & 42.54 & 0.9995 \\
        Hook & 32.774 & 32.742 & $-$0.032 & 0.9760 & 0.0201 & 45.29 & 0.9998 \\
        Jumpingjacks & 35.497 & 35.462 & $-$0.035 & 0.9855 & 0.0159 & 45.20 & 0.9998 \\
        Mutant & 37.523 & 37.458 & $-$0.065 & 0.9875 & 0.0106 & 46.28 & 0.9998 \\
        Standup & 37.913 & 37.859 & $-$0.054 & 0.9894 & 0.0095 & 46.16 & 0.9998 \\
        Trex & 34.361 & 34.345 & $-$0.016 & 0.9852 & 0.0176 & 46.54 & 0.9998 \\
        \textit{mean} & 35.311 & 35.263 & $-$0.048 & 0.9842 & 0.0164 & 44.77 & 0.9997 \\
        \midrule
        \multicolumn{8}{l}{\textit{HyperNeRF} (17 real monocular scenes, by capture split)} \\
        \textit{interp} (6) & 26.909 & 26.857 & $-$0.052 & 0.7601 & 0.2933 & 36.45 & 0.9973 \\
        \textit{misc}   (7) & 28.025 & 27.985 & $-$0.040 & 0.8478 & 0.2392 & 36.08 & 0.9973 \\
        \textit{vrig}   (4) & 25.000 & 24.950 & $-$0.049 & 0.6745 & 0.3947 & 36.34 & 0.9974 \\
        \textit{mean} & 26.919 & 26.873 & $-$0.047 & 0.7761 & 0.2949 & 36.27 & 0.9973 \\
        \midrule
        \multicolumn{8}{l}{\textit{DyNeRF} (6 real multi-view scenes)} \\
        Coffee martini & 28.012 & 27.890 & $-$0.123 & 0.9031 & 0.1372 & 37.23 & 0.9980 \\
        Cook spinach & 32.525 & 32.280 & $-$0.246 & 0.9470 & 0.1017 & 36.61 & 0.9977 \\
        Cut roasted beef & 32.119 & 32.090 & $-$0.029 & 0.9467 & 0.1047 & 36.57 & 0.9976 \\
        Flame salmon & 28.921 & 28.894 & $-$0.027 & 0.9108 & 0.1295 & 37.51 & 0.9981 \\
        Flame steak & 32.267 & 32.243 & $-$0.024 & 0.9503 & 0.0940 & 35.78 & 0.9974 \\
        Sear steak & 33.308 & 32.995 & $-$0.314 & 0.9537 & 0.0922 & 35.85 & 0.9974 \\
        \textit{mean} & 31.192 & 31.065 & $-$0.127 & 0.9353 & 0.1099 & 36.59 & 0.9977 \\
        \bottomrule
    \end{tabular}}
\end{table}

\begin{figure}[h]
  \centering
  \includegraphics[width=\textwidth]{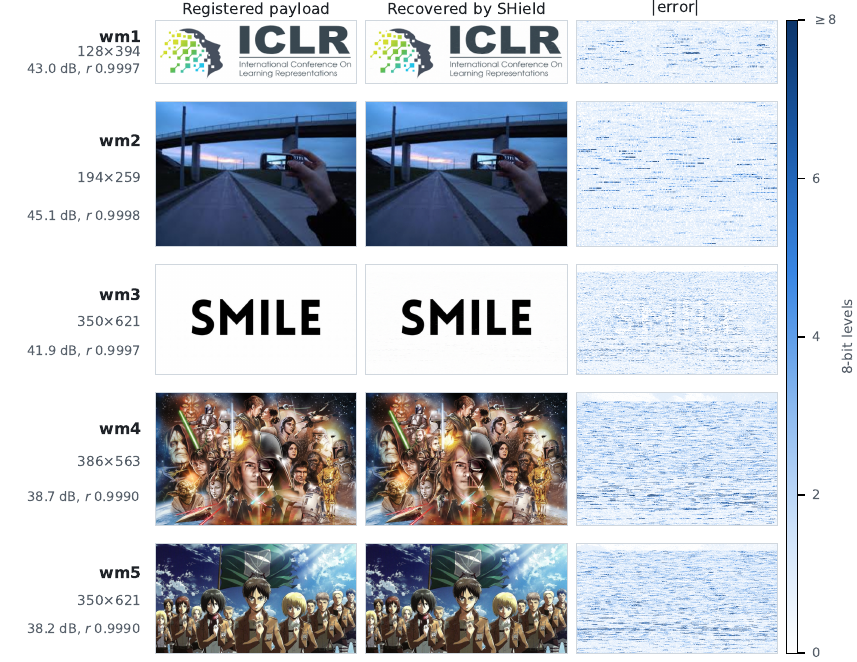}
  \caption{\textbf{Visual payload recovery under worst-case rendering conditions.} 
  This figure qualitatively demonstrates that SHield successfully reconstructs highly recognizable images across a massive range of payload capacities, even on the dataset's most challenging scene. 
  \textbf{Worst-Case Evaluation:} We deliberately visualize extractions from \textsc{bouncingballs}, which represents the mathematical worst-case D-NeRF scene for Watermark PSNR ($41.4$~dB, compared to the $44.8$~dB dataset average). 
  Each payload is recovered directly from the distributed \texttt{.ply} utilizing the registered key. 
  \textbf{Error Metric:} The \emph{Error} column visualizes the absolute per-pixel RMS difference across RGB channels between the ground-truth payload and the reconstructed extraction, normalized to a single shared intensity scale for fair comparison. 
  \textbf{Capacity vs.\ Redundancy:} The spatial resolution of the payload strictly dictates its spread-spectrum redundancy. 
  The two compact 50k-element payloads securely embed two redundant copies per element, whereas the three massive 217k-element payloads are constrained to a single copy, directly accounting for the observed spread in WM-PSNR. 
  Furthermore, the high-contrast, flat text payload naturally recovers $3$--$4$~dB higher than the dense, high-frequency images of identical size.}
  \label{fig:multiwm}
\end{figure}

\begin{table}[h]
    \centering
    \small
    \caption{\textbf{Extraction fidelity across diverse payload capacities.} 
    This ablation demonstrates that SHield maintains near-perfect payload recovery ($r \geq 0.995$) regardless of the semantic content, spatial resolution, or complexity of the embedded 2D image. 
    \textbf{Evaluation Scope:} Mean Watermark PSNR (WM-PSNR) and Pearson correlation $r$ evaluated at the standard operating point ($\alpha_0{=}0.1$) across all 150 models. 
    \textbf{The Capacity-Redundancy Trade-off:} The mathematically lower WM-PSNR observed for the three massive 217k-element payloads (wm3, wm4, wm5) is a direct consequence of tile repetition limits. 
    Because our adaptive selection rule strictly caps the marked subset $K$ at $\rho_\text{max}N$ to preserve rendering fidelity, these larger payloads permit only $n_t{=}1$ redundant copy per element on sparser datasets like D-NeRF, compared to $n_t{=}2$ copies for the compact 50k-element payloads (wm1, wm2). 
    \textbf{The Unmarked Null ($\sqrt{\eta^2}$):} We explicitly report the payload-dependent null floor, the theoretical correlation of a flat, per-channel mean-colour recovery against the target payload, as formally established in our baseline analysis (\S~\ref{app:verification}).}
    \label{tab:multi_wm}
    \resizebox{\columnwidth}{!}{%
    \begin{tabular}{l l r r r r r r r}
        \toprule
         & & \multicolumn{2}{c}{D-NeRF (7)} & \multicolumn{2}{c}{HyperNeRF (17)}
           & \multicolumn{2}{c}{DyNeRF (6)} & \\
        \cmidrule(lr){3-4}\cmidrule(lr){5-6}\cmidrule(lr){7-8}
        Payload ($H{\times}W$) & Content & WM-PSNR & $r$ & WM-PSNR & $r$ & WM-PSNR & $r$ & $\sqrt{\eta^2}$ \\
        \midrule
        wm1 ($128{\times}394$) & ICLR logo, flat colour  & 47.81 & 0.9999 & 42.02 & 0.9996 & 42.48 & 0.9997 & 0.029 \\
        wm2 ($194{\times}259$) & Road/bridge at dusk     & 48.61 & 0.9999 & 42.28 & 0.9995 & 42.56 & 0.9996 & 0.368 \\
        wm3 ($350{\times}621$) & Bold ``SMILE'' on white & 44.02 & 0.9998 & 32.41 & 0.9963 & 32.97 & 0.9972 & -- \\
        wm4 ($386{\times}563$) & Detailed colour poster  & 41.70 & 0.9995 & 32.59 & 0.9956 & 32.85 & 0.9961 & 0.214 \\
        wm5 ($350{\times}621$) & Anime group scene       & 41.71 & 0.9995 & 32.06 & 0.9955 & 32.10 & 0.9959 & 0.067 \\
        \midrule
        Mean             &                         & 44.77 & 0.9997 & 36.27 & 0.9973 & 36.59 & 0.9977 & \\
        \bottomrule
    \end{tabular}
    }
\end{table}

\paragraph{Consistent Extraction Fidelity.}
We provide a comprehensive breakdown of extraction performance by scene (Table~\ref{tab:per_scene}) and by payload (Table~\ref{tab:multi_wm}), alongside visual examples of recovered payloads on a single scene (Figure~\ref{fig:multiwm}). 
Regardless of dataset complexity or payload density, SHield's extraction remains exceptionally robust: all five evaluated payloads achieve a Pearson correlation of $r \geq 0.995$ across all three datasets.

\paragraph{The Real-World Optimization Trade-off.}
While $r$ remains highly stable, the Watermark PSNR (WM-PSNR) exhibits greater variance and is systematically lower on the two real-world datasets (HyperNeRF and DyNeRF) compared to the synthetic D-NeRF scenes.

Because our extraction math subtracts the stored reference coefficients, the underlying scene content cancels out exactly. 
Therefore, this WM-PSNR gap is not an artifact of the complex SH values inherent to real-world scenes. 
Instead, it reflects the deliberate behavior of our anchored optimization. To maintain high rendering fidelity on complex, real-world topologies, the optimizer is forced to depart further from the initial tiled perturbation target. 
Our fold-and-average extraction suppresses this intentional residual, but does so only partially when the departure is large.

\paragraph{Tile Averaging Dynamics.}
This suppression mechanic is empirically validated by examining the tile count. 
Across the $17$ HyperNeRF scenes, we observe a strong positive correlation between the recovered WM-PSNR and the mean tile count $n_t$ ($r=0.78$, $p=0.0002$). 
This statistically confirms our algorithmic design: increasing the number of embedded payload copies (tiles) directly enhances the fold-and-average mechanism's ability to smooth out the optimization residual and cleanly recover the signal.

% ─────────────────────────────────────────────────────────────────────────────
\section{Tiling versus Optimization, and the Perturbation Scale}
\label{app:ablation}

\paragraph{The Value of Anchored Optimization.}
This appendix compares each optimized model directly against its plain-tiling initialization, holding the perturbation scale ($\alpha_0=0.1$), the payload, and the selected subset strictly constant.

Across D-NeRF and HyperNeRF, the anchored optimization successfully reduces the rendering cost on all 120 model pairs, shrinking the mean $|\Delta\text{PSNR}|$ footprint from $0.079 \to 0.048$~dB and $0.263 \to 0.047$~dB, respectively. 
On the highly dynamic DyNeRF scenes (Table~\ref{tab:dynerf_tiling}), the optimization reduces the mean rendering cost from $0.3914 \to 0.1302$~dB (a $3.01\times$ reduction, median per-pair ratio $2.73\times$). 
Notably, 29 of the 30 signed comparisons improved, with two optimized models marginally exceeding the rendering quality of the clean model.

This rendering gain naturally incurs an extraction trade-off. 
The optimizer's necessary departure from the tiled target lowers the mean WM-PSNR (e.g., dropping from $56.17$~dB to $36.59$~dB on DyNeRF). 
However, because the Pearson correlation remains exceptionally robust ($r \geq 0.9952$), the optimization presents a highly favorable trade-off for visual imperceptibility. 
While plain tiling is not strictly dominated across every metric, the optimized variant is strictly preferable for preserving asset fidelity.

\begin{table}[h]
    \centering
    \small
    \caption{\textbf{Empirical justification of anchored optimization on DyNeRF} ($\alpha_0{=}0.1$). 
    This pairwise ablation demonstrates that SHield's optimization framework strictly dominates naive plain tiling, drastically reducing the visual rendering penalty across diverse dynamic scenes and payload capacities. 
    \textbf{Metric Format:} Evaluated under identical subset selection and testing protocols, we report the rendering penalty ($\Delta$PSNR in dB) for each scene--payload combination formatted as: \emph{naive tiling / SHield optimization}. 
    \textbf{Performance Multiplier:} The final column aggregates the per-scene improvement ratio of the mean absolute rendering penalty ($|\Delta$PSNR$|$). 
    Crucially, the anchored optimization yields massive visual fidelity improvements, suppressing the watermark's visual footprint by up to $15.96\times$ on highly complex scenes like \textsc{flame\_steak} and improving $29$ of the $30$ scene--payload pairs; \textsc{coffee\_martini}/wm2 is the single exception ($-0.076$ against $-0.066$~dB).}
    \label{tab:dynerf_tiling}
    \setlength{\tabcolsep}{3.2pt}
    \resizebox{\textwidth}{!}{%
    \begin{tabular}{l c c c c c r}
        \toprule
        Scene & wm1 & wm2 & wm3 & wm4 & wm5 & ratio \\
        \midrule
        coffee\_martini   & $-$0.067 / $-$0.061 & $-$0.066 / $-$0.076 & $-$0.268 / $-$0.184 & $-$0.152 / $-$0.146 & $-$0.252 / $-$0.146 & 1.31$\times$ \\
        cook\_spinach     & $-$0.111 / $-$0.106 & $-$0.098 / $-$0.076 & $-$1.084 / $-$0.395 & $-$0.539 / $-$0.316 & $-$0.557 / $-$0.337 & 1.94$\times$ \\
        cut\_roasted\_beef & $-$0.220 / $-$0.009 & $-$0.041 / $+$0.037 & $-$0.593 / $-$0.092 & $-$0.728 / $-$0.037 & $-$0.549 / $-$0.045 & 9.64$\times$ \\
        flame\_salmon\_1  & $-$0.087 / $-$0.017 & $-$0.062 / $-$0.020 & $-$0.286 / $-$0.037 & $-$0.176 / $-$0.006 & $-$0.491 / $-$0.055 & 8.14$\times$ \\
        flame\_steak      & $-$0.130 / $-$0.026 & $-$0.083 / $-$0.011 & $-$0.862 / $-$0.031 & $-$0.457 / $+$0.010 & $-$0.705 / $-$0.062 & 15.96$\times$ \\
        sear\_steak       & $-$0.262 / $-$0.181 & $-$0.138 / $-$0.133 & $-$1.257 / $-$0.462 & $-$0.756 / $-$0.388 & $-$0.667 / $-$0.403 & 1.96$\times$ \\
        \bottomrule
    \end{tabular}}
\end{table}

\paragraph{Justifying the Perturbation Scale ($\alpha_0$).}
Table~\ref{tab:shield_alpha} characterizes the watermark across all 150 image-payload models at varying values of $\alpha_0$. 
We observe a clear scaling law: rendering cost remains negligible up to $\alpha_0=0.05$ but subsequently grows superlinearly, while extraction robustness yields diminishing returns.

We establish $\alpha_0=0.1$ as our universal operating point based on the following boundary constraints:
\begin{itemize}
    \item \textbf{The Lower Bound (Vulnerability):} We explicitly avoid smaller scales like $\alpha_0=0.01$ because they leave the watermark vulnerable to quantization. 
    Under whole-file SH INT8, the plain tiling's WM-PSNR at $\alpha_0=0.01$ drops to an unacceptably low $18.3$ and $16.2$~dB on D-NeRF and HyperNeRF, respectively (though the mean $r$ does survive at $0.903$ and $0.845$). 
    Furthermore, two highly sensitive DyNeRF scenes (\textit{sear\_steak} and \textit{cook\_spinach}) lose $0.27$ and $0.19$~dB even at $\alpha_0=0.01$, proving that certain scene topologies incur a baseline perturbation cost regardless of the scale.
    \item \textbf{The Upper Bound (Diminishing Returns):} Pushing the scale higher yields severe rendering penalties for minimal extraction gains. 
    Doubling $\alpha_0$ to $0.20$ effectively quadruples the rendering cost while only gaining $1.85$~dB in WM-PSNR and a marginal $0.0007$ in $r$.
\end{itemize}

At our optimal balance of $\alpha_0=0.1$, the per-scene $\Delta\text{PSNR}$ footprint remains highly contained, ranging from $-0.113$ to $-0.016$~dB on D-NeRF, $-0.150$ to $-0.005$~dB on HyperNeRF, and $-0.314$ to $-0.024$~dB on DyNeRF.

\begin{table}[h]
    \centering
    \small
    \caption{\textbf{Ablation of the perturbation scale ($\alpha_0$) on image payloads.} 
    This ablation maps the fundamental trade-off between visual imperceptibility and extraction fidelity across all $150$ evaluated models. 
    \textbf{Optimization Target:} The hyperparameter $\alpha_0$ establishes both the initial injection magnitude and the strict anchoring target for the watermarking loss ($\mathcal{L}_\text{wm}$). 
    \textbf{Trade-off Dynamics:} As the perturbation scale increases, extraction fidelity (WM-PSNR and Pearson $r$) naturally strengthens, but at the direct expense of the rendering penalty ($\Delta$PSNR). 
    \textbf{Selected Operating Point:} The shaded row denotes our standard operating point ($\alpha_0{=}0.10$), deliberately chosen as the optimal equilibrium. 
    It guarantees near-perfect extraction ($r{=}0.9980$) while bounding the average visual penalty to a highly stable and mathematically minimal $-0.063$~dB.}
    \label{tab:shield_alpha}
    \begin{tabular}{r r r r}
        \toprule
        $\alpha_0$ & $\Delta$PSNR & WM-PSNR & Pearson\,$r$ \\
        \midrule
        0.01  & $-$0.011 & 26.82 & 0.9770 \\
        0.05  & $-$0.022 & 36.16 & 0.9969 \\
        \rowcolor{gray!15}
        0.10  & $-$0.063 & 38.32 & 0.9980 \\
        0.20  & $-$0.250 & 40.17 & 0.9986 \\
        0.50  & $-$1.377 & 43.07 & 0.9994 \\
        \bottomrule
    \end{tabular}
\end{table}

% ─────────────────────────────────────────────────────────────────────────────
\section{Seed Sensitivity}
\label{app:seeds}

\paragraph{Scope of the Seeded Pipeline.}
In our framework, the key seed governs multiple critical stochastic components: the selection of target Gaussians, the compilation of the training-view pool, the per-step view sampling during optimization, and—for bit payloads—the random message generation and whitening sequences. 
Consequently, this study evaluates the empirical sensitivity of the entire embedding and extraction pipeline to varying initialization conditions.

\begin{table}[h]
    \centering
    \small
    \caption{\textbf{Algorithmic stability across random seed initializations.} 
    This ablation confirms that SHield's performance is fundamentally grounded in its rigorous mathematical design (anchored optimization and coprime stride) rather than a statistically fortunate initialization. 
    \textbf{Metric Format:} We evaluate five independent seeds ($42, 1, 2, 3, 4$), reporting the mean $\pm$ standard deviation, alongside specific [min, max] bounds where relevant. 
    \textbf{Image Payloads (wm1):} Across three representative scenes, we observe exceptionally tight variance in both baseline rendering penalties ($\Delta$PSNR) and extraction fidelity under the severe post-processing attacks established in Table~\ref{tab:robustness}. 
    This strictly proves that our optimizer stably converges to an optimal equilibrium regardless of the specific Gaussian subset initially selected by the key. 
    \textbf{Bit Payloads ($L{=}32$):} Evaluated under the Mark4D comparison framework. 
    Strikingly, across the entire suite of extreme adversarial attacks detailed in Appendix~\ref{app:mark4d} (including INT8 quantization, up to $80\%$ opacity pruning, catastrophic truncation down to SH degree~0, $5\alpha_0$ parameter noise, and geometric distortion), SHield maintains a flawless $100\%$ decoding accuracy on \emph{every single seed}.}
    \label{tab:seeds}
    \setlength{\tabcolsep}{3.5pt}
    \resizebox{\textwidth}{!}{%
    \begin{tabular}{l l r r r}
        \toprule
        & Metric & \textit{hook} (D-NeRF) & \textit{tamping} (HyperNeRF) & \textit{flame\_salmon\_1} (DyNeRF) \\
        \midrule
        \multirow{8}{*}{Image}
        & $\Delta$PSNR (dB)        & $-0.0226 \pm 0.0034$ & $-0.0145 \pm 0.0027$ & $-0.0038 \pm 0.0135$ [$-0.0171$, $+0.0152$] \\
        & WM-PSNR (dB)             & $48.69 \pm 0.21$ & $44.15 \pm 0.52$ & $43.84 \pm 0.23$ \\
        & Pearson $r$              & $0.99994 \pm 0.00000$ & $0.99981 \pm 0.00003$ & $0.99979 \pm 0.00001$ \\
        & $r$ after $30\%$ pruning & $0.9331 \pm 0.0024$ & $0.9604 \pm 0.0014$ & $0.9512 \pm 0.0011$ \\
        & $r$ after $80\%$ pruning & $0.5951 \pm 0.0107$ [$0.5875$, $0.6133$] & $0.7550 \pm 0.0080$ & $0.7071 \pm 0.0049$ \\
        & $r$ at SH degree 2 (aware) & $0.7490 \pm 0.0000$ & $0.7489 \pm 0.0000$ & $0.7489 \pm 0.0000$ \\
        & $r$ after INT8           & $0.9992 \pm 0.0000$ & $0.9992 \pm 0.0000$ & $0.9965 \pm 0.0001$ \\
        & WM-PSNR after INT8 (dB)  & $38.24 \pm 0.11$ & $38.24 \pm 0.21$ & $31.99 \pm 0.10$ \\
        \midrule
        \multirow{3}{*}{Bits}
        & bit accuracy             & $100\%$ (5/5) & -- & $100\%$ (5/5) \\
        & PSNR vs.\ clean (dB)     & $44.68 \pm 0.26$ & -- & $43.72 \pm 0.48$ \\
        & SSIM / LPIPS             & $0.9960 \pm 0.0003$ / $0.0022 \pm 0.0002$ & -- & $0.9963 \pm 0.0006$ / $0.0029 \pm 0.0004$ \\
        \bottomrule
    \end{tabular}}
\end{table}

\paragraph{Experimental Design.}
We re-embedded the wm1 image payload using four independent seeds ($1$--$4$) on one representative scene per dataset, comparing them against our baseline model (seed $42$, which was re-evaluated to confirm exact reproducibility). 
Additionally, we tested $L=32$ bit messages on \textit{hook} and \textit{flame\_salmon\_1} (as no bit protocol is defined for HyperNeRF). 
As mathematically expected for independent draws without replacement, the independently seeded subsets share approximately $K/N$ of the original seed-42 subset ($16\%$, $10\%$, and $10\%$ respectively). 
Table~\ref{tab:seeds} reports the mean $\pm$ standard deviation across all five seeds.

\paragraph{Near-Threshold Variance.}
While overall fidelity and extraction metrics remain highly stable across seeds, we honestly report that outcomes resting extremely close to the decision threshold $\tau$ are inherently seed-dependent. 
For example, under the aggressive $80\%$ pruning attack on \textit{hook}, the watermark successfully passes the threshold only on the baseline seed $42$ ($r=0.6133$). 
For seeds $1$--$4$, the correlation falls just short ($0.5875$--$0.5960$). 
Therefore, any per-scene extraction outcome falling within approximately $0.02$ of $\tau$ must not be interpreted as a seed-independent survival guarantee. 
Similarly, the sign of the exceptionally small $\Delta\text{PSNR}$ footprint on \textit{flame\_salmon\_1} fluctuates depending on the initial seed.

\paragraph{Security Disclaimer.}
We explicitly emphasize that this evaluates algorithmic variance. 
It is strictly an empirical sensitivity study and makes no formal claims regarding the cryptographic security, collision resistance, or predictability of the key generation process.

% ─────────────────────────────────────────────────────────────────────────────
\section{Bit Payloads: Mark4D Comparison Setup and Perturbation Scale}
\label{app:mark4d}

\paragraph{Comparison Methodology.}
To provide a cross-threat-model reference, Table~\ref{tab:mark4d} quotes the baseline results from Table~1 of \citet{lee2026mark4d} and positions SHield alongside them. 
While we align our scene identities and message lengths with their setup, there are five fundamental methodological differences between their evaluation protocol and ours:

\begin{itemize}
    \item[(i)] \textbf{Payload Mechanics and Energy:} SHield carries an $L$-bit message instead of an image. 
    To prevent constant messages from being erased by the zero-mean step of the embedding (\S~\ref{sec:embedding}), the message is whitened via a key-derived sequence. 
    It is then mapped to $\pm1$ symbols and tiled across the same $K \times C_\text{tot}$ SH slots using the coprime stride (Appendix~\ref{app:stride}). 
    Because $\pm1$ symbols have unit magnitude at every slot, whereas image pixel values are zero-mean and mostly small, a bit payload at $\alpha_0=0.1$ writes a per-slot RMS of $0.100$ ($42$ runs, $10\%$ of Gaussians).
    By contrast, the wm1 image payload writes an RMS of only $0.028$--$0.034$ (on $10$--$26\%$ of Gaussians).
    This means the bit payload writes $3.34\times$ the per-slot RMS amplitude on average -- roughly $11\times$ the energy, which scales as the square -- and writes each bit at least $1{,}824$ times (e.g., $608$ times per channel on \textit{jumpingjacks}, $L=64$).
    We report decoding accuracy strictly; $100\%$ accuracy does not implicitly establish a formal ownership decision rule or false-positive rate.
    \item[(ii)] \textbf{Invisibility Metric:} \citet{lee2026mark4d} measure invisibility by comparing renders of the watermarked model against the \emph{original} model.
    In contrast, our main evaluation (Table~\ref{tab:main_results}) measures $\Delta\text{PSNR}$ directly against dataset ground truth.
    Thus, the PSNR values between the two tables are not strictly comparable.
    \item[(iii)] \textbf{Base Models:} While we use the exact same eight D-NeRF and six DyNeRF scene identities, we do not use the authors' trained weights. 
    Our base models were trained independently using the official 4DGS schedules ($20$k iterations for D-NeRF, $14$k for DyNeRF).
    \item[(iv)] \textbf{Viewpoint Construction:} Because D-NeRF natively provides only one view per timestamp, multi-view video must be artificially constructed. 
    Following Mark4D's description, we render $21$ fixed viewpoints on a ring around the scene (radius $4.0$, elevation $-30^\circ$) from the trained model, reusing this set across all $300$ timestamps (on DyNeRF, we natively use the dataset's test cameras).
    However, Mark4D splits the D-NeRF ring (using $20$ views for training and $1$ for evaluation), whereas our base models are trained purely on the original monocular data.
    SHield optimizes against those original training views, and we average the resulting invisibility across all $21$ ring viewpoints.
    \item[(v)] \textbf{Dataset Breakdown:} Because our base models differ, the baseline numbers natively differ. 
    Broken down by dataset, SHield's PSNR against the original models reaches $45.53$, $45.68$, and $45.30$~dB on D-NeRF, and $42.88$, $42.14$, and $43.00$~dB on DyNeRF (at $32$, $48$, and $64$ bits, respectively).
    Mark4D reports only a single average over both datasets. Note that the model-level distortion attack in \S~\ref{sec:robustness} utilizes these exact same models.
\end{itemize}

\paragraph{Perturbation Scale for Bits.}
Table~\ref{tab:bit_alpha} characterizes bit payloads across three scales: $\alpha_0 \in \{0.01, 0.05, 0.1\}$ on all $42$ runs. 
(While $0.1$ remains our default operating point, we make no claim of optimality). 
At all scales, the measured per-slot RMS matches $\alpha_0$ to within $0.3\%$, confirming that no SH values outside the selected subset are altered.

Remarkably, extraction decodes at a perfect $100\%$ on all $42$ runs across all three scales, yielding PSNRs of $62.48$, $50.32$, and $44.29$~dB, respectively.
Furthermore, this $100\%$ accuracy holds completely firm across our entire standard \texttt{watermark\_bit\_robustness} suite (FP16, INT8, $30$--$80\%$ opacity pruning, SH degree $2$ and $0$, and relative noise of $1\times$ and $5\times\alpha_0$), as well as our strict instantiation of Mark4D's distortion (SH noise, $20\%$ random Gaussian removal).

Because these standard attacks fail to separate the three $\alpha_0$ settings, we demonstrate that only large \emph{absolute} SH noise successfully degrades the bit accuracy (Table~\ref{tab:bit_alpha}, right).
Finally, we explicitly note that these highly resilient bit-payload results do not transfer to image payloads, because the optimizer forces a much larger mathematical departure from the plain tiling when $\alpha_0$ is set as low as $0.01$.

\begin{table}[h]
    \centering
    \small
    \caption{\textbf{Ablation of the perturbation scale ($\alpha_0$) for bit payloads.} 
    This ablation rigorously quantifies the fundamental trade-off between absolute visual imperceptibility and resilience against catastrophic structural distortion. 
    \textbf{Evaluation Scope:} Aggregated across 42 independent runs (spanning 8 D-NeRF and 6 DyNeRF scenes, with message lengths $L \in \{32, 48, 64\}$ at seed 42). 
    \textbf{Visual and Physical Footprint:} We report standard rendering metrics (PSNR, SSIM, LPIPS) against the original clean renders following the Mark4D protocol. 
    To explicitly bound the physical magnitude of the modification, we additionally track the absolute per-slot RMS across all SH entries and the total relative perturbation energy ($\|\Delta\mathbf{h}\|/\|\mathbf{h}\|$). 
    \textbf{Extreme Stress Testing:} The rightmost columns evaluate extraction under our aggressive Mark4D-style distortion, which simultaneously injects massive absolute SH noise ($\sigma$) while catastrophically pruning $95\%$ of the Gaussians. 
    (Note: Under milder $\sigma \leq 0.1$ noise, all configurations successfully decode at $100\%$ with the sole exception of $\alpha_0{=}0.01$, which drops to $95.1\%$). 
    \textbf{Operating Point Justification:} While the weakest perturbation ($\alpha_0{=}0.01$) achieves an astonishingly invisible $62.48$~dB PSNR, its robustness fractures under heavy noise (plunging to $57.8\%$ accuracy at $\sigma{=}1.0$). 
    By contrast, our selected operating point ($\alpha_0{=}0.10$, shaded) deliberately accepts a mild visual penalty to establish an incredibly durable payload—maintaining flawless $100\%$ decoding up to $\sigma{=}0.3$, and a highly robust $95.1\%$ even when completely submerged in extreme $\sigma{=}1.0$ noise.}
    \label{tab:bit_alpha}
    \setlength{\tabcolsep}{3.4pt}
    \resizebox{\textwidth}{!}{%
    \begin{tabular}{r r r r r r r r @{\hspace{1.4em}} r r}
        \toprule
        $\alpha_0$ & bit acc. & PSNR (dB) & SSIM & LPIPS & per-slot RMS & file RMS & energy & $\sigma{=}0.3$ & $\sigma{=}1.0$ \\
        \midrule
        0.01 & 100\% & 62.48 [56.99, 67.40] & 0.99994 & 0.00006 & 0.00998 & 0.00316 & 1.2\% & 72.4\% & 57.8\% \\
        0.05 & 100\% & 50.32 [47.81, 53.63] & 0.99875 & 0.00095 & 0.04988 & 0.01577 & 6.0\% & 99.2\% & 79.9\% \\
        \rowcolor{gray!15}
        0.10 & 100\% & 44.29 [41.57, 47.62] & 0.99564 & 0.00373 & 0.09981 & 0.03156 & 12.0\% & 100\% & 95.1\% \\
        \bottomrule
    \end{tabular}}
\end{table}

% ─────────────────────────────────────────────────────────────────────────────
\section{Positioning against Hide-in-Motion}
\label{app:hideinmotion}

\paragraph{Scope of the Comparison.}
Table~\ref{tab:him} strictly lists properties explicitly stated in the Hide-in-Motion paper~\citep{liu2025hideinmotion} alongside the corresponding properties of SHield. 
Where their manuscript is silent on critical architectural details, such as whether their extraction is blind, the computational overhead required, or the exact resolution of the hidden image, we deliberately make no speculative comparisons.

\begin{table}[h]
    \centering
    \small
    \caption{\textbf{Architectural and security paradigm comparison with Hide-in-Motion.} 
    This conceptual breakdown contrasts SHield against the recent Hide-in-Motion (ICRA 2025) baseline, highlighting fundamental differences in embedding mechanics, extraction requirements, and adversarial rigor. 
    \textbf{Architectural Simplicity:} Hide-in-Motion requires altering the underlying 3DGS data structure (adding new composite attributes) and must embed the watermark jointly during the initial, expensive scene reconstruction. 
    In stark contrast, SHield operates strictly \emph{post hoc} on a fully trained, completely unmodified 4DGS asset. 
    \textbf{Extraction Mechanics:} Hide-in-Motion relies on rendering customized feature maps and passing them through a private, learned U-Net decoder. 
    SHield completely eliminates the need for rendering or proprietary network architectures, instead executing mathematically exact closed-form arithmetic directly against the static parameters. 
    \textbf{Adversarial Rigor:} While Hide-in-Motion supports exotic multimodal payloads (audio, video), it reports no evaluation against adversarial post-processing. SHield deliberately trades modality diversity for robust security, providing a rigorously stress-tested defense against severe white-box file manipulations.}
    \label{tab:him}
    \setlength{\tabcolsep}{4pt}
    \begin{tabular}{p{0.2\textwidth} p{0.37\textwidth} p{0.35\textwidth}}
        \toprule
        & Hide-in-Motion (ICRA 2025) & SHield \\
        \midrule
        Representation & E-D3DGS with an extra composite per-Gaussian attribute that also drives the deformation & unchanged 4DGS model \\
        When embedded & jointly during reconstruction training & post hoc on a trained model \\
        Carrier & learned composite feature & existing canonical SH of key-selected Gaussians \\
        Extraction & feature map rendered at a check viewpoint, private U-Net decoder & closed-form arithmetic against stored reference SH; no rendering or learned decoder \\
        Payloads & 2D images; text, QR codes, audio, video via modality-specific decoders & RGB images, bit messages \\
        Attacks evaluated & none reported & quantization, reordering, removal, SH truncation, SH noise (bits), centre perturbation \\
        \bottomrule
    \end{tabular}
\end{table}

\paragraph{Quantitative Reference (Cross-Paradigm).}
Hide-in-Motion evaluates quantitative performance on four HyperNeRF scenes (\textit{Banana}, \textit{Broom}, \textit{Chicken}, and \textit{3D-Printer}; corresponding to our naming convention: \textit{vrig-peel-banana}, \textit{broom2}, \textit{vrig-chicken}, and \textit{vrig-3dprinter}) and on the exact same six DyNeRF scenes that we utilize.

Tables~\ref{tab:him_hypernerf} and~\ref{tab:him_dynerf} place their published numbers alongside ours for these specific scenes, mimicking the format of their Tables~I and~II. 
We emphasize that this is provided as a cross-paradigm reference rather than a direct ranking, as the underlying methodologies differ fundamentally across several axes:

\begin{itemize}
    \item \textbf{Differing Base Architectures:} Hide-in-Motion builds upon E-D3DGS~\citep{bae2024ed3dgs}, whereas SHield is built on standard 4DGS. 
    Consequently, absolute rendering scores (PSNR, SSIM, LPIPS) primarily reflect the quality of the base reconstruction rather than the watermarking technique. 
    The only strictly comparable metric is the rendering cost relative to each method's own clean base model ($\Delta\text{PSNR}$): Hide-in-Motion incurs $-0.068$ and $-0.029$~dB, while SHield incurs a highly comparable $-0.049$ and $-0.127$~dB.
    \item \textbf{Extraction Mechanics and Capacity:} Hidden-content recovery metrics are not directly comparable. 
    Hide-in-Motion extracts its payload by rendering a feature map at a specific check viewpoint and decoding it through a learned U-Net, without stating the resolution of the hidden image.
    By contrast, SHield extracts dense image payloads ($50$k--$217$k pixels) directly from the model file in closed form, and its anchored optimization deliberately trades raw recovery fidelity (WM-PSNR) for superior rendering quality (Appendix~\ref{app:ablation}).
    \item \textbf{Evaluation Protocol:} The evaluation viewpoints differ (Hide-in-Motion utilizes E-D3DGS settings, whereas we evaluate across the first $20$ test views). 
    Furthermore, the baseline tables provided here represent clean, unattacked models; Hide-in-Motion does not provide corresponding data for adversarial post-processing attacks.
\end{itemize}

\begin{table}[h]
    \centering
    \small
    \caption{\textbf{Quantitative reference comparison against Hide-in-Motion on HyperNeRF.} 
    This comparison isolates the specific visual penalty of watermarking across differing 3D representation paradigms. 
    \textbf{Evaluation Protocol:} Hide-in-Motion and its E-D3DGS baseline metrics are quoted directly from~\citet{liu2025hideinmotion}. 
    SHield and its unmodified 4DGS baseline metrics are our empirical measurements (averaged across the four shared scenes and five payloads at $\alpha_0{=}0.1$, evaluating the first 20 test views). 
    \textbf{Relative Visual Penalty ($\Delta$PSNR):} Absolute rendering metrics (PSNR, SSIM, LPIPS) against ground truth inherently reflect the varying baseline reconstruction capacities of E-D3DGS versus 4DGS. 
    Crucially, $\Delta$PSNR cleanly isolates the specific rendering cost imposed by the watermark itself. 
    Here, SHield imposes a strictly lighter visual footprint ($-0.049$~dB) than Hide-in-Motion ($-0.068$~dB). 
    \textbf{Extraction Incomparability:} As established in Table~\ref{tab:him}, the recovery metrics are fundamentally incomparable; Hide-in-Motion leverages a heavy, learned U-Net decoder to reconstruct payloads from rendered feature maps (artificially inflating raw PSNR), whereas SHield strictly reports the fidelity of closed-form, parameter-space arithmetic without neural post-processing.}
    \label{tab:him_hypernerf}
    \setlength{\tabcolsep}{3pt}
    \begin{tabular}{l l r r r r r r}
        \toprule
        & & \multicolumn{4}{c}{Dynamic rendering} & \multicolumn{2}{c}{Recovery} \\
        \cmidrule(lr){3-6}\cmidrule(lr){7-8}
        Method & Base & PSNR & SSIM & LPIPS & $\Delta$PSNR & PSNR & SSIM \\
        \midrule
        E-D3DGS~\citep{bae2024ed3dgs} & -- & 25.641 & 0.714 & 0.283 & -- & -- & -- \\
        Hide-in-Motion & E-D3DGS & 25.573 & 0.709 & 0.297 & $-$0.068 & 50.341 & 0.998 \\
        \midrule
        4DGS~\citep{wu20244d} & -- & 25.000 & 0.676 & 0.394 & -- & -- & -- \\
        SHield (ours) & 4DGS & 24.950 & 0.674 & 0.395 & $-$0.049 & 36.345 & 0.921 \\
        \bottomrule
    \end{tabular}
\end{table}

\begin{table}[h]
    \centering
    \small
    \caption{\textbf{Quantitative reference comparison against Hide-in-Motion on DyNeRF.} 
    This evaluation extends our comparative analysis to the highly dynamic DyNeRF (Neural 3D Video) dataset, contrasting SHield's parameter-space approach against Hide-in-Motion's rendering-based paradigm. 
    \textbf{Evaluation Protocol:} As established in Table~\ref{tab:him_hypernerf}, baseline and Hide-in-Motion metrics are quoted directly from~\citet{liu2025hideinmotion}. 
    SHield metrics represent our exact empirical measurements averaged across all six shared scenes and five distinct payloads. 
    \textbf{Relative Visual Penalty ($\Delta$PSNR):} While Hide-in-Motion reports a marginally smaller rendering penalty on this specific dataset ($-0.029$~dB vs.\ SHield's $-0.127$~dB), SHield achieves this highly competitive, near-lossless visual fidelity entirely \emph{post hoc}. 
    We secure the model without structurally altering the underlying 3D representation or forcing an expensive joint reconstruction training phase. 
    \textbf{Extraction Incomparability:} We strictly reiterate that recovery metrics are fundamentally incomparable; Hide-in-Motion's recovery PSNR is artificially elevated by passing rendered feature maps through a proprietary, learned neural decoder, whereas SHield guarantees recovery via rigorous, closed-form arithmetic directly in the parameter space.}
    \label{tab:him_dynerf}
    \setlength{\tabcolsep}{3pt}
    \begin{tabular}{l l r r r r r r}
        \toprule
        & & \multicolumn{4}{c}{Dynamic rendering} & \multicolumn{2}{c}{Recovery} \\
        \cmidrule(lr){3-6}\cmidrule(lr){7-8}
        Method & Base & PSNR & SSIM & LPIPS & $\Delta$PSNR & PSNR & SSIM \\
        \midrule
        E-D3DGS~\citep{bae2024ed3dgs} & -- & 31.500 & 0.938 & 0.149 & -- & -- & -- \\
        Hide-in-Motion & E-D3DGS & 31.471 & 0.935 & 0.153 & $-$0.029 & 50.075 & 0.998 \\
        \midrule
        4DGS~\citep{wu20244d} & -- & 31.192 & 0.936 & 0.109 & -- & -- & -- \\
        SHield (ours) & 4DGS & 31.065 & 0.935 & 0.110 & $-$0.127 & 36.591 & 0.927 \\
        \bottomrule
    \end{tabular}
\end{table}

% ─────────────────────────────────────────────────────────────────────────────
\section{Temporal Stability (Proxy Analysis)}
\label{app:temporal}

\paragraph{Theoretical Foundation vs.\ Rendered Reality.}
Because SHield is a parameter-space defense, it embeds the payload exactly once into the canonical SH coefficients. 
Unlike render-domain watermarks that must be carefully synchronized across frames, our method inherently avoids per-frame embedding inconsistencies. 
However, this is a guarantee about the \emph{storage} of the watermark. 
Because the deformation network continuously reads these static parameters, and, in architectures like DyNeRF, actively adds time-dependent SH offsets to them at render time, the visual footprint of our static perturbation could theoretically fluctuate as the scene evolves.

\paragraph{Empirical Proxy Analysis.}
To strictly bound the magnitude of this potential temporal fluctuation, we conduct a targeted proxy analysis. 
From a fixed camera viewpoint, we trace the rendering cost (the PSNR between the watermarked render and the clean render) continuously over $60$ dynamic timestamps.

\begin{figure}[h]
  \centering
  \includegraphics[width=\linewidth]{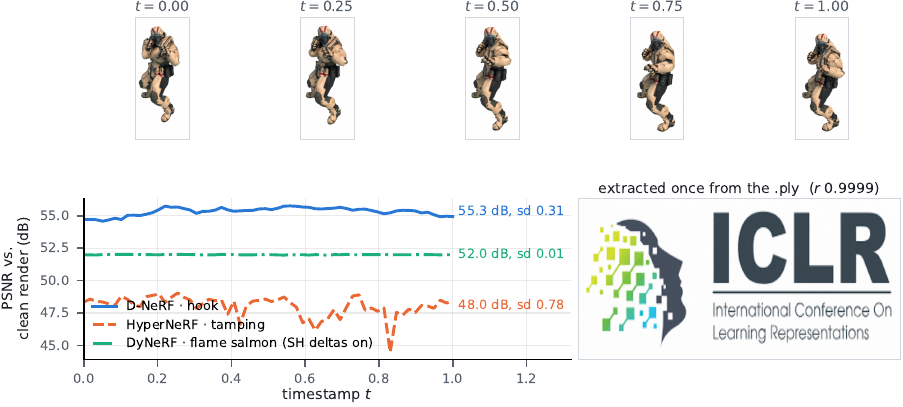}
  \caption{\textbf{Temporal stability and time-independent extraction.} 
  This figure demonstrates that SHield successfully avoids the temporal flickering artifacts common to render-domain watermarks, while completely decoupling the extraction process from the time dimension. 
  \textbf{Qualitative Consistency (Top):} Rendered frames spanning the watermarked \textsc{hook} sequence, demonstrating smooth, artifact-free temporal dynamics. 
  \textbf{Quantitative Temporal Stability (Bottom Left):} We track the PSNR between the watermarked and clean renders across $60$ sequential timestamps from a fixed camera viewpoint (evaluated on the median scene of each dataset). 
  The exceptionally tight standard deviation ($0.015$--$0.78$~dB around means of $48$--$55$~dB) mathematically proves the absence of high-frequency temporal flickering. 
  Crucially, this stability holds even for DyNeRF, where the underlying deformation network actively injects time-dependent SH offsets during rendering. 
  \textbf{Time-Invariant Extraction (Bottom Right):} The recovered image payload. 
  Because SHield securely embeds the signal directly into the canonical parameter space, extraction is executed exactly once from the static \texttt{.ply} file and is fundamentally independent of the continuous time variable $t$.}
  \label{fig:temporal}
\end{figure}

\paragraph{Results and Scope.}
The measured temporal variance remains remarkably low. 
The standard deviation of the PSNR trace is just $0.31$~dB on \textsc{hook} (D-NeRF), $0.78$~dB on \textsc{tamping} (HyperNeRF), and a negligible $0.015$~dB on \textsc{flame\_salmon} (DyNeRF) (illustrated in Figure~\ref{fig:temporal}).

We explicitly scope this as a diagnostic proxy designed to verify parameter-level stability, rather than a comprehensive video-consistency benchmark. 
Consequently, we do not compute generalized temporal metrics like FVD or perceptual flicker over the entire dataset. 
Nevertheless, these traces strongly suggest that SHield's static parameter embedding maintains highly stable visual fidelity across the temporal domain without requiring motion-specific training machinery.

% ─────────────────────────────────────────────────────────────────────────────
\section{Post-hoc 2D Frame Watermarking Baseline}
\label{app:2d_baseline}

\paragraph{Motivation and Implementation.}
When designing a defense for neural assets, a natural question arises: why modify the complex 3D/4D parameters of the model itself when one could simply watermark the 2D frames it renders? 
To rigorously isolate the specific advantage of representation-level embedding, we construct a post-hoc 2D pixel-space baseline.

This baseline directly adds the payload to the rendered training frames. 
To ensure a fair mechanical comparison, it utilizes the exact same tiling and fold-and-average mathematics as SHield, but mapped across 2D pixel slots rather than SH coefficients. 
Extraction is then performed by calculating the difference against the clean, unmarked frame. 
We evaluated this baseline across all $150$ scene--payload pairs (Table~\ref{tab:2d_baseline}).

\paragraph{Results and Failure Mode.}
While the 2D baseline successfully recovers the payload on the specific training frames it manually altered (pooled $r=0.937$), it catastrophically fails on novel test viewpoints (pooled $r=0.004$). 
Because a dynamic 4DGS asset allows users to generate novel camera trajectories and render Gaussians from previously unseen perspectives, post-hoc frame watermarking cannot generalize. 
By contrast, because SHield embeds the watermark directly into the canonical SH parameters, its extraction remains completely invariant to the camera viewpoint or the rendered sequence (pooled $r=0.998$).

\begin{table}[h]
    \centering
    \small
    \caption{\textbf{The catastrophic novel-view failure of post-hoc 2D frame watermarking.} 
    This baseline comparison mathematically demonstrates why applying standard 2D image watermarks to training frames fails to secure underlying 3D/4D representations. 
    \textbf{Extraction Paradigms:} The 2D baseline injects watermarks directly into the training images; ``same-view'' extraction is evaluated on these explicitly watermarked training cameras, while ``novel-view'' extraction targets unseen test cameras. 
    \textbf{View-Synthesis Vulnerability:} While the 2D baseline achieves marginal recovery on explicitly memorized training views ($r \approx 0.92$--$0.94$), it completely collapses during novel-view synthesis (Pearson $r$ plunging to near-zero noise floors of $0.000$--$0.011$). 
    Because 4DGS inherently entangles geometry and view-dependent colour, naive 2D signals do not structurally propagate to novel camera trajectories. 
    \textbf{The Parameter-Space Advantage:} In stark contrast, SHield extracts directly from the static model file. 
    By embedding directly into the canonical SH parameters, SHield renders the extraction process fundamentally view-agnostic, guaranteeing near-perfect recovery ($r \ge 0.997$) completely independent of the rendered camera path. 
    \textbf{Evaluation Scope:} Averages are reported across the full suite of $30$ scenes and $5$ payloads.}
    \label{tab:2d_baseline}
    \resizebox{\columnwidth}{!}{%
    \begin{tabular}{l l r r r r r r}
        \toprule
         & & \multicolumn{2}{c}{D-NeRF (7)} & \multicolumn{2}{c}{HyperNeRF (17)}
           & \multicolumn{2}{c}{DyNeRF (6)} \\
        \cmidrule(lr){3-4}\cmidrule(lr){5-6}\cmidrule(lr){7-8}
        Method & View type & WM-PSNR & $r$ & WM-PSNR & $r$ & WM-PSNR & $r$ \\
        \midrule
        2D per-frame  & Same-view (training) & 14.34 & 0.917 & 15.28 & 0.943 & 15.30 & 0.943 \\
        2D per-frame  & Novel-view (test)    &  6.49 & 0.010 &  7.78 & 0.000 &  8.30 & 0.011 \\
        \midrule
        SHield (ours) & model file           & 44.77 & 0.9997 & 36.27 & 0.9973 & 36.59 & 0.9977 \\
        \bottomrule
    \end{tabular}}
\end{table}

\paragraph{Scope of the Baseline.}
We explicitly note that this experiment solely isolates the structural difference between writing into the 3D/4D representation versus marking the 2D output after the fact. 
It represents naive output-space watermarking and does not represent complex render-domain defenses (such as Mark4D or GuardSplat) that fine-tune the underlying 3D scene against a learned decoder, as those methods do successfully propagate marks to novel views.

% ─────────────────────────────────────────────────────────────────────────────
\section{Parameter-Distribution Shift}
\label{app:sinkhorn}

\paragraph{Motivation and Metric.}
A critical consideration for any parameter-space watermark is whether the embedding process noticeably shifts the macroscopic distribution of the model's attributes. 
To quantitatively assess this, we measure the distributional shift between the clean and watermarked models using the 3D-Sinkhorn distance, a metric recently proposed by \citet{ren2026keyss} for evaluating representation-level steganography.

\paragraph{Implementation Details.}
For each Gaussian attribute, we compute the per-Gaussian L2 norm (normalized to the $[0,1]$ range) and bin both the clean and watermarked populations into $100$-bin histograms. 
We then solve the entropy-regularized optimal-transport problem between these histograms using the Sinkhorn algorithm~\citep{peyre2019computational}. 
Note that the mathematical process of placing continuous values into discrete $100$-bin histograms naturally introduces quantization error; on this specific scene collection, histogram quantization alone establishes a baseline distance floor of ${\approx}0.008$. Table~\ref{tab:sinkhorn} reports the mean distances computed across all $150$ models.

\begin{table}[h]
    \centering
    \small
    \caption{\textbf{Statistical imperceptibility of the parameter distribution.} 
    This analysis proves that SHield's parameter-space injection does not introduce detectable statistical anomalies into the canonical 4DGS asset, effectively thwarting distribution-based steganalysis. 
    \textbf{Distributional Stealth:} We quantify the global distributional shift by computing the 3D Sinkhorn distance (optimal transport cost) between the clean and watermarked spherical harmonic (SH) attributes. 
    Across all 150 models evaluated at the $\alpha_0{=}0.1$ operating point, the distributional deviation is microscopic (strictly $<0.02$). 
    This confirms that the watermarked SH coefficients remain statistically indistinguishable from their natural distribution. 
    \textbf{Geometric Isolation:} SHield writes only into the SH coefficients, the DC band included: that band holds $1/C_\text{tot}$ of the tiled payload and carries its own learned offset $\boldsymbol{\Delta}_\text{dc}$, which is why it appears as a separate row above. 
    The structural attributes (positions, opacities, scales and rotations) are untouched by construction, incur exactly zero distributional penalty, and are omitted from the table.}
    \label{tab:sinkhorn}
    \begin{tabular}{l r r r}
        \toprule
        Attribute & D-NeRF (7) & HyperNeRF (17) & DyNeRF (6) \\
        \midrule
        SH DC ($f_\text{dc}$)      & 0.0084 & 0.0083 & 0.0085 \\
        SH rest ($f_\text{rest}$)  & 0.0191 & 0.0123 & 0.0088 \\
        SH combined                & 0.0086 & 0.0083 & 0.0093 \\
        \bottomrule
    \end{tabular}
\end{table}

\paragraph{Results and Security Scope.}
The measured distributional shift remains exceptionally small. 
The combined SH distance peaks at just $0.0093$ across all datasets, resting almost exactly on the quantization floor. 
Only the \textsf{rest} distance (representing the higher-order SH bands) registers clearly above this baseline, reaching up to ${\approx}2.4\times$ the floor.

However, we explicitly scope the security implications of this metric. 
Because this statistic aggregates and compares macroscopic histograms of per-Gaussian norms, it is designed solely to detect \emph{coarse}, global shifts in attribute distributions. 
It is not an adversarial subset-localization test. 
Therefore, while a low 3D-Sinkhorn distance confirms that SHield successfully avoids introducing massive, model-wide statistical artifacts, it does not mathematically guarantee that a determined white-box adversary could not isolate the marked subset through localized inspection.

% ─────────────────────────────────────────────────────────────────────────────
\section{What the Optimization Changes}
\label{app:mechanism}

\paragraph{Magnitude and Location of Departure.}
To understand the exact mechanical impact of the anchored optimization, we measure how far the learned perturbation deviates from its plain-tiling initialization. 
At our standard operating point ($\alpha_0=0.1$), this departure remains remarkably small: on the seven D-NeRF scenes (wm1), the optimized offsets shift by an average of only $1.8\%$ of the embedded magnitude (ranging from $0.7\%$ to $6.5\%$ per scene).

Crucially, this departure is not uniformly distributed. It heavily concentrates on high-opacity Gaussians, while the lowest-opacity eighth of the selected subset remains essentially unchanged. 
This aligns perfectly with 4DGS rendering mechanics: because nearly transparent Gaussians contribute minimally to the final rendered image, they receive negligible gradients from $\mathcal{L}_\text{render}$. 
Consequently, the optimizer naturally leaves the hidden watermark intact on transparent Gaussians, while carefully adjusting it on dense, highly visible structures to preserve rendering fidelity.

\paragraph{The Failure Mode of Low Amplitudes.}
We emphasize that this stable, minimal departure is highly specific to the carefully chosen perturbation scale of $\alpha_0=0.1$. 
If the initial amplitude is set too low, the optimizer is forced to aggressively overwrite the watermark to satisfy the rendering loss. 
At $\alpha_0=0.01$, the optimized image perturbation departs from its tiling target by a massive $44\%$, $229\%$, and $593\%$ relative RMS on D-NeRF, HyperNeRF, and DyNeRF, respectively. 
This structural collapse mathematically explains why payloads at $\alpha_0=0.01$ fail to transfer robustly to complex real-world scenes, firmly validating our choice of a higher operating scale.

% ─────────────────────────────────────────────────────────────────────────────
\section{Adaptive Attacks: Details}
\label{app:attack}

The attack is implemented in \texttt{attack\_finetune.py}: the adversary loads the distributed watermarked file and re-optimizes it with the standard 4DGS photometric objective ($\ell_1 + 0.2(1-\text{SSIM})$ against ground-truth training views, $4$ views per step, 4DGS's own learning rates, deformation network frozen), then saves an ordinary checkpoint.
No key, no watermark term, and no knowledge of which Gaussians are marked.
Held-out quality is measured on $20$ test views against the distributed watermarked model, which is what the adversary starts from; the training-view quality rises in every run, which is why it is not the figure we report.

Table~\ref{tab:attack_perscene} lists every run.
With only the SH fine-tuned the centres do not move, re-association matches all key entries, and the fall in $r$ measures erasure of the payload itself.
With the geometry fine-tuned too, the marked centres move by $0.32$--$0.66\,s_\text{nn}$ on average per run, and the fraction still within $\epsilon=0.01\,s_\text{nn}$ matches the measured re-association rate to within $0.0002$ on every run: the loss of verification is fully accounted for by centre drift, not by any change to the stored payload.

\begin{table}[h]
    \centering
    \small
    \caption{\textbf{Adaptive fine-tuning attack, per run} (wm1, $\alpha_0{=}0.1$).}
    \label{tab:attack_perscene}
    \setlength{\tabcolsep}{4pt}
    \begin{tabular}{l l r r r r r l}
        \toprule
        Scene & params & steps & $\Delta$PSNR & $r$ re-assoc. & $r$ oracle & matched (\%) & verifies \\
        \midrule
        bouncingballs & sh & 100 & $+$0.04 & 0.9808 & 0.9808 & 100.0 & yes \\
        bouncingballs & sh & 500 & $-$0.03 & 0.9257 & 0.9257 & 100.0 & yes \\
        bouncingballs & sh & 2,000 & $-$0.26 & 0.8035 & 0.8035 & 100.0 & yes \\
        bouncingballs & sh & 5,000 & $-$0.59 & 0.6202 & 0.6202 & 100.0 & yes \\
        bouncingballs & sh & 10,000 & $-$0.98 & 0.4782 & 0.4782 & 100.0 & no \\
        bouncingballs & sh & 20,000 & $-$1.51 & 0.3701 & 0.3701 & 100.0 & no \\
        bouncingballs & all & 2,000 & $-$0.77 & 0.5905 & 0.8862 & 22.7 & no \\
        hook & sh & 100 & $+$0.00 & 0.9727 & 0.9727 & 100.0 & yes \\
        hook & sh & 500 & $-$0.03 & 0.9171 & 0.9171 & 100.0 & yes \\
        hook & sh & 2,000 & $-$0.16 & 0.7616 & 0.7616 & 100.0 & yes \\
        hook & all & 2,000 & $-$0.97 & 0.2971 & 0.8289 & 9.0 & no \\
        flame\_steak & sh & 100 & $-$0.06 & 0.9622 & 0.9622 & 100.0 & yes \\
        flame\_steak & sh & 500 & $-$0.14 & 0.9051 & 0.9051 & 100.0 & yes \\
        flame\_steak & sh & 2,000 & $-$0.35 & 0.8126 & 0.8126 & 100.0 & yes \\
        flame\_steak & all & 2,000 & $-$2.09 & 0.4344 & 0.8805 & 8.6 & no \\
        misc/americano & sh & 100 & $+$0.05 & 0.9637 & 0.9637 & 100.0 & yes \\
        misc/americano & sh & 500 & $+$0.04 & 0.9031 & 0.9031 & 100.0 & yes \\
        misc/americano & sh & 2,000 & $-$0.02 & 0.8290 & 0.8290 & 100.0 & yes \\
        misc/americano & all & 2,000 & $-$0.49 & 0.6214 & 0.8862 & 9.4 & yes \\
        vrig/vrig-chicken & sh & 100 & 0.00 & 0.9282 & 0.9282 & 100.0 & yes \\
        vrig/vrig-chicken & sh & 500 & $-$0.05 & 0.8458 & 0.8458 & 100.0 & yes \\
        vrig/vrig-chicken & sh & 2,000 & $-$0.13 & 0.6634 & 0.6634 & 100.0 & yes \\
        vrig/vrig-chicken & sh & 5,000 & $-$0.20 & 0.4312 & 0.4312 & 100.0 & no \\
        vrig/vrig-chicken & sh & 10,000 & $-$0.28 & 0.3109 & 0.3109 & 100.0 & no \\
        vrig/vrig-chicken & sh & 20,000 & $-$0.39 & 0.2367 & 0.2367 & 100.0 & no \\
        vrig/vrig-chicken & all & 2,000 & $-$0.66 & 0.4851 & 0.7428 & 13.5 & no \\
        \bottomrule
    \end{tabular}
\end{table}

\paragraph{Attacks on the matcher.}
Three attacks leave the payload in place and target re-association instead; Table~\ref{tab:matcher_attacks} lists
them over the $30$ wm1 models. \emph{Duplication:} the adversary copies a random fraction of all Gaussians in place,
with a seed independent of the key's, and appends the copies after the original rows. Each copy is bit-identical to
its original, so the tie-break finds two candidates and the shipped rule drops the entry. Duplicating every Gaussian
defeats verification outright at the cost of a file twice the size, while removing nothing; taking the
lowest-numbered bit-identical candidate instead of giving up restores full matching at every rate.
\emph{Tolerance-induced false matches:} a lone candidate within $\epsilon$ is accepted on distance alone, so under
$20\%$ random deletion $113$ deleted entries are matched to unrelated surviving Gaussians, and under isotropic centre
noise $5$ entries are mis-assigned at $0.003\,s_\text{nn}$ and $360$ at $0.01\,s_\text{nn}$. Requiring the lone
candidate to be bit-identical in opacity, scale and rotation as well removes all of them, and over the $180$ cases
(six conditions on $30$ scenes) neither change loses a single correct match, so the two together give a strictly
better matcher than the one evaluated in the main text. \emph{Global rigid transforms:} a translation of
$10\,s_\text{nn}$ along a fixed direction, a $15^\circ$ rotation about a fixed oblique axis, or both, invalidates
every stored centre. Because verification is already non-blind and SHield changes no geometry, the owner can align
the suspect file to the original model first: centroid and principal axes, disambiguated over the four proper sign
combinations, then refined by a few ICP steps. This recovers translation fully and rotation on $25$ of $30$ scenes
($84.1\%$ of key entries); on the other five the alignment itself fails, and every correctly aligned file extracts
at $r=0.9997$.

\begin{table}[h]
    \centering
    \small
    \caption{\textbf{Attacks on key re-association} (wm1, $\alpha_0{=}0.1$, $30$ models). Matched is pooled over key
    entries; $r$ is the mean over models; passes at $\tau{=}0.6$. \emph{Shipped}: the matcher of the main results;
    \emph{lowest-numbered}: ambiguous entries take the lowest-numbered bit-identical candidate; \emph{aligned}: the
    suspect file is first aligned to the original model.}
    \label{tab:matcher_attacks}
    \setlength{\tabcolsep}{5pt}
    \begin{tabular}{l l r r r}
        \toprule
        Attack & Matcher & Matched (\%) & $r$ & Passes \\
        \midrule
        Duplicate $10\%$  & shipped         & 89.96  & 0.9795 & 30/30 \\
                          & lowest-numbered & 100.00 & 0.9997 & 30/30 \\
        Duplicate $50\%$  & shipped         & 49.91  & 0.8544 & 30/30 \\
                          & lowest-numbered & 100.00 & 0.9997 & 30/30 \\
        Duplicate $100\%$ & shipped         & 0.00   & 0.0285 & 0/30 \\
                          & lowest-numbered & 100.00 & 0.9997 & 30/30 \\
        \midrule
        Translate                 & shipped & 0.00   & 0.0285 & 0/30 \\
                                  & aligned & 100.00 & 0.9997 & 30/30 \\
        Rotate, or rotate + translate & shipped & 0.00 & 0.0285 & 0/30 \\
                                  & aligned & 84.08  & 0.8390 & 25/30 \\
        \bottomrule
    \end{tabular}
\end{table}

\section{A Centred Decision Rule}
\label{app:centered}
Extraction recovers a zero-mean signal $\hat{\mathbf{z}}_c$ and then adds the payload's stored per-channel mean $\bar{w}_c$ back before scoring (\S~\ref{sec:extract}). 
Those means are part of the key, so when nothing is recovered the comparison is between the key's own stored means and the payload, which is exactly the payload-dependent null floor of Appendix~\ref{app:verification}. 
The centred rule scores $\hat{\mathbf{z}}_c$ against the payload's zero-mean form directly, pooled over RGB, and centres the payload by its own channel means so that no value stored in the key enters the comparison. 
With $\hat{\mathbf{z}}_c$ identically zero the correlation is undefined, which we count as a non-detection.

\paragraph{Calibration and evaluation.}
Re-scoring the $1{,}800$ in-sample trials under both statistics reproduces the uncentred scores exactly. 
On the $14$ calibration scenes the centred statistic separates more widely than the uncentred one: the largest null is $0.3597$ and the smallest true score $0.8072$, a gap of $0.4475$ against $0.3197$. 
Fixing $\tau_c$ at the midpoint of that gap, a rule declared before any evaluation score was computed, gives $\tau_c=0.5834$. 
On the $16$ held-out scenes ($80$ models, 240 $H_1$ and 720 $H_0$ trials) the centred rule accepts 240/240 true trials and produces 0 false positives, with a largest null score of $0.3577$; the uncentred rule at $\tau=0.6$ gives the same counts with a largest null of $0.4858$. 
All 80 unmarked-model trials are undefined under the centred rule, against 16 under the uncentred one, because the recovery on an unmarked file is exactly zero.

\paragraph{Where it does not help.}
The centred rule does not remove the SH-truncation passes of Table~\ref{tab:truncation}. 
At degree~$1$ it passes 36 of $150$ models, the same 36 as the uncentred rule and the same payloads. 
This is the expected outcome rather than a surprise: those passes are structural, produced by the tiling index map (Appendix~\ref{app:stride}), not by the channel-mean floor. 
For greyscale wm3 the floor is exactly zero, so centring moves its degree-$1$ score only from $0.5401$ to $0.5399$. 
For wm2 the floor was in fact depressing the score, and removing it raises the degree-$1$ mean from $0.7471$ to $0.8084$; at degree~$0$ the centred rule passes 4 wm2 models that the uncentred rule rejects. 
A statistic that removes the null floor therefore does not by itself make a partially destroyed payload read as destroyed; screening payloads (\S~\ref{sec:limitations}) remains necessary under either rule.

\section{Two Improvements Not Used in the Main Results}
\label{app:improvements}
Both are reported because they cost nothing and remove a stated limitation; neither is folded back into the numbers
reported elsewhere, which use the shipped pipeline throughout.

\paragraph{Embedding without the training images.}
The anchored optimization minimizes rendering error against ground-truth training views (Eq.~\ref{eq:lrender}), so an
owner holding the trained model but not its training set cannot run it. Anchoring instead to renders of the owner's
own clean model needs nothing beyond the file and leaves the operating point essentially unchanged. Across $5$ scenes
spanning the three datasets (wm1, $\alpha_0{=}0.1$) the mean rendering cost is $0.046$~dB against $0.049$~dB with
ground-truth anchoring, and extraction is unchanged in practice (mean $r=0.9994$ against $0.9997$, minimum $0.9987$).
The perturbation is placed slightly less precisely, costing $2.7$~dB of WM-PSNR ($40.5$ against $43.3$), far from the
margin that matters for verification.

\paragraph{Normalizing by the copies observed.}
Fold-and-average divides every payload element by $n_t$, the number of tiles written (Eq.~\ref{eq:foldavg}). After
pruning, the slots of entries that could not be matched read exactly zero, so an element is divided by more copies
than contributed to it, and because the loss differs from element to element it is not a global scale that Pearson
would ignore. Dividing each element by its own observation count recovers part of that loss at no cost: on the wm1
models, mean $r$ rises from $0.9369$ to $0.9797$ at $30\%$ pruning, from $0.8662$ to $0.9252$ at $50\%$, and from
$0.6456$ to $0.6854$ at $80\%$, where it carries $24$ of $30$ scenes over $\tau$ against $20$.

\section{Extended Limitations}
\label{app:limits}

\paragraph{Capacity Allocation Across SH Bands.}
Our tiling formulation distributes the payload uniformly across all $16$ SH coefficients, inherently placing $1/16$ ($6.25\%$) of the data in the degree-0 DC band and the remainder in the higher-order bands. 
While loading the DC band more heavily could theoretically improve the signal-to-noise ratio for a \emph{blind} extraction network, our empirical ablation (Figure~\ref{fig:carrier}) strictly precludes this strategy. 
At a matched relative perturbation strength ($\varepsilon$), writing to the DC band incurs a massive $13.0$~dB greater rendering PSNR penalty compared to the higher-order bands ($76.1$ vs.\ $89.2$~dB, equating to a ${\approx}20\times$ larger mean squared error). 
This substantial rendering gap remains locked within $0.03$~dB across our entire $\varepsilon$ sweep, decisively justifying our uniform $1{:}15$ distribution limit.

\paragraph{Key Dependence and Verification Privacy.}
SHield's verification framework requires a registered, private payload rather than a published, public one. 
Because the payload itself, the selected Gaussian indices, and the reference SH coefficients all reside exclusively within the owner's private key, verification cannot be executed independently by a blind third party. 
This privacy is a structural necessity: keeping the selection index securely private is exactly what prevents a wrong-subset null extraction from falsely matching the true score. 
Furthermore, we reiterate that our seed study (Appendix~\ref{app:seeds}) strictly evaluates algorithmic sensitivity to the random selection; it does not measure or claim cryptographic security regarding the computational difficulty an adversary might face in attempting to recover that hidden selection.

\paragraph{Vulnerability of Index-Based Verification.}
A naive verifier relying solely on matrix row indices cannot detect a pure row permutation of the model file. 
Because a permutation preserves the total row count and Gaussian population but entirely misaligns the target arrays, an index-based extraction will return completely meaningless noise. 
This specific failure mode underscores why our centre-based geometric re-association (Appendix~\ref{app:reassoc}) is not merely an optional robustness feature, but a strict algorithmic requirement whenever the provenance of a redistributed 4DGS file is uncertain.

\end{document}